\documentclass{article} % For LaTeX2e

\PassOptionsToPackage{table}{xcolor}
\usepackage{colm2024_conference}
\usepackage[table]{xcolor}
\usepackage{colortbl}

\usepackage[utf8]{inputenc} % allow utf-8 input
\usepackage[T1]{fontenc}    % use 8-bit T1 fonts
\usepackage{url}            % simple URL typesetting
\usepackage{booktabs}       % professional-quality tables (只保留一个)
\usepackage{amsfonts}       % blackboard math symbols (只保留一个)
\usepackage{amsmath}
\usepackage{amssymb}
\usepackage{nicefrac}       % compact symbols for 1/2, etc.
\usepackage{microtype}      % microtypography
\usepackage{graphicx}

\usepackage{algorithm}
\usepackage{algpseudocode}  % (只保留一个)

\usepackage{multirow}       % (只保留一个)
\usepackage{makecell}
\usepackage{enumitem}
\usepackage{multicol}
\usepackage{array}

\usepackage{wrapfig}
\usepackage{subcaption}
\usepackage[most]{tcolorbox} % (只保留一个)
\usepackage{soul}
\usepackage{pifont}

\usepackage{todonotes}

\usepackage{hyperref}       % hyperlinks (只保留一个，放在最后)

\definecolor{deltaBg}{RGB}{220,230,255} % subtle blue-grey
\newcommand{\normalrow}{\rowcolor{gray!30}}
\newcommand{\rowhighlight}{\rowcolor{deltaBg}}

\tcbset{
  aibox/.style={
    top=10pt,
    colback=white,
    colframe=black,
    colbacktitle=black,
    enhanced,
    center,
    attach boxed title to top left={yshift=-0.1in,xshift=0.15in},
    boxed title style={boxrule=0pt,colframe=white,},
  }
}
\newtcolorbox{AIbox}[2][]{aibox, title=#2,#1}

\newcommand{\framework}{Embodied Planner-R1}

\title{Unleashing Embodied Task Planning Ability in LLMs via Reinforcement Learning}

\colmfinalcopy

\author{
Zhaoye Fei$^{1}$, \hspace{.1em} Li Ji$^{1}$,\hspace{.1em} Siyin Wang$^{1,2}$,\hspace{.1em} Junhao Shi$^{1,2}$\\
\textbf{Jingjing Gong$^{2\dagger}$, \hspace{.1em} and Xipeng Qiu$^{1,2\dagger}$} \\
$^1$Fudan University $^2$Shanghai Innovation Institute\\
\texttt{zyfei20@fudan.edu.cn, jil24@m.fudan.edu.cn},  \\
\texttt{jjgongjj@gmail.com, xpqiu@fudan.edu.cn }
}

\fancypagestyle{firstpage}{
  \lhead{\begin{picture}(0,0)\put(0,-6){\includegraphics[width=0.3\linewidth]{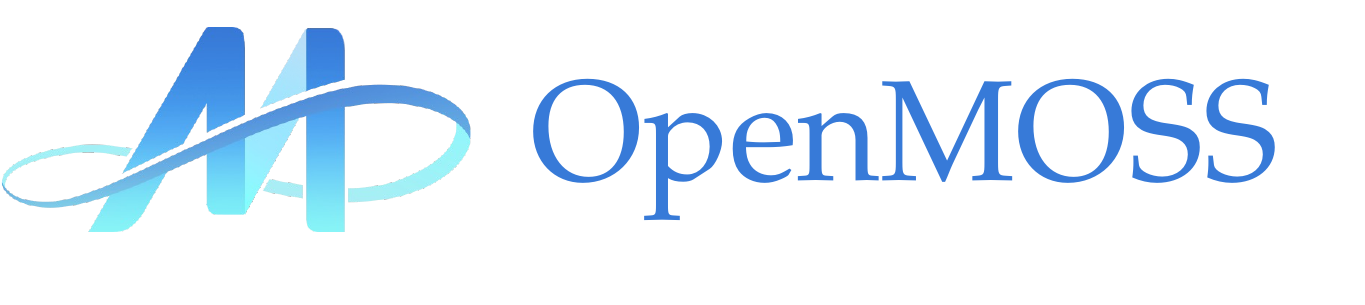}}\end{picture}}
}

\begin{document}

\maketitle

\thispagestyle{firstpage}

\newcommand\blfootnote[1]{%
\begingroup
\renewcommand\thefootnote{}\footnote{#1}%
\addtocounter{footnote}{-1}%
\endgroup
}

\blfootnote{$^\dagger$ Corresponding authors.}
\blfootnote{$\cdot$ We will release our models and code on \url{https://github.com/OpenMOSS/Embodied-Planner-R1} to support future research in interactive planning.}

\begin{abstract}\label{sec:abs}
Large Language Models (LLMs) have demonstrated remarkable capabilities across various tasks, yet they face significant challenges in embodied task planning scenarios that require continuous environmental understanding and action generation. Existing approaches generate open-loop action scripts based on static knowledge, making it difficult to learn causal relationships between actions and environmental feedback, particularly in partially observable environments. We introduce~\textbf{\framework}, a novel outcome-driven reinforcement learning framework that enables LLMs to develop interactive capabilities through autonomous exploration with minimal supervision. Our framework incorporates three key innovations: (1) Without human annotations, we employ pure reinforcement learning with group rollout, incorporating in-environment interaction through parallel exploration; (2) completion-driven sparse reward; and (3) Interactive Policy Optimization (IPO) for efficient learning from grouped trajectories. Across two challenging text-based Embodied planning benchmarks, \framework\ achieves impressive completion rates of ~\textbf{97.78\%} on ALFWorld and ~\textbf{79.92\%} on ScienceWorld, surpassing prior methods by a large margin, and suffers only a -3.66\% drop in previously unseen environments, evidencing strong generalization. 
\end{abstract}

\section{Introduction}\label{sec:intro}

\begin{wrapfigure}{r}{0.45\textwidth}
    \raggedleft
    \vspace{-7em}
    \resizebox{0.35\textwidth}{!}{
    \includegraphics[]{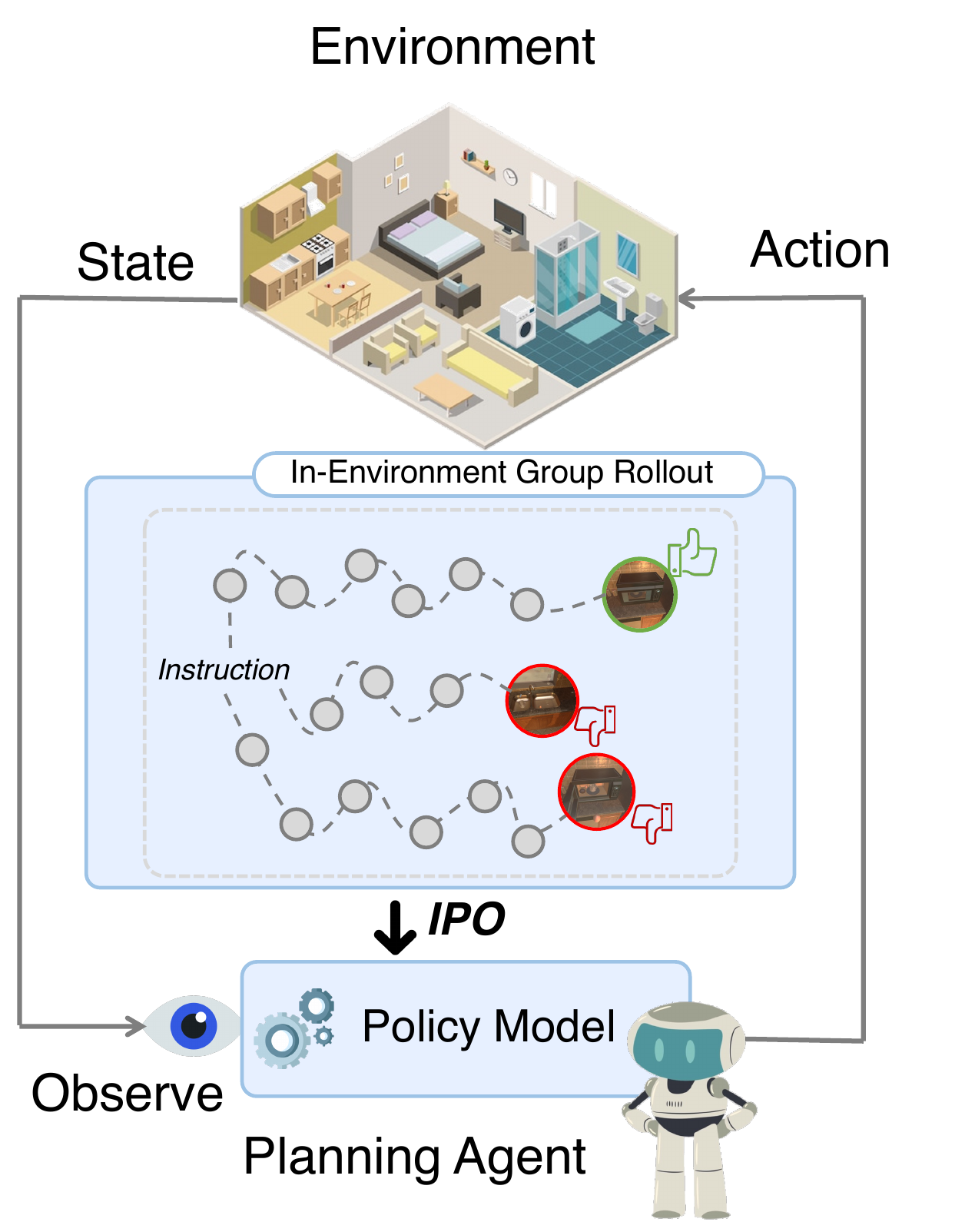}}
    \caption{The overview of \framework, which encourages agents to explore environments autonomously, training with binary rewards based on task completion status.}
    \label{fig:overview}
    \vspace{-2em}              
\end{wrapfigure}

Large Language Models (LLMs) have demonstrated remarkable capabilities across a wide range of tasks, from open-domain dialogue to complex reasoning~\citep{tom20gpt3,openai2023gpt4,guo2024deepseekr1,shao2024deepseekmath,cai24internlm2,yang2024qwen2_5}. Nonetheless, translating these abilities into effective planning in interactive environments remains an open challenge. Planning agents must continually perceive the world, reason about it, and act, forming a tightly coupled decision-making loop. Current LLMs still lack the situational interactive awareness needed for the interactive planning task. In contrast, by actively probing the world, human learn how their actions transform the environment and refine both our perception and world comprehension~\citep{Gibson1979GIBTEA}. Such exploration-driven learning, while intuitive for us, poses a substantial challenge for artificial intelligence, particularly when applying LLMs to interactive planning domains like robotics and embodied agents.

Traditional LLM-based planners treat the model as a static decision maker~\citep{chen2023robogpt, min2022film}, they generate a fixed action sequence from pre-trained knowledge and ignore real-time feedback. Although~\citet{yao2023react} improved task performance by introducing reasoning processes during planning, these methods still amount to offline rollouts based on prior knowledge; they fail to establish the crucial causal link between actions and environmental feedback during interaction. This shortcoming is fatal in partially observable or unfamiliar environments, where an agent must continually revise its world model to succeed. Recent work~\citep{song2024eto, wang2024nat, chen2024fireact, zeng2024agenttuning} has attempted to close the gap with additional training by incorporating human-curated demonstration sets or handcrafted knowledge. However, the heavy reliance on human priors limits the LLM's scalability and constrains its overall capability. 
%Thus, a fundamental problem emerges: how can we guide an LLM with interaction intelligence acquired through minimally supervised, autonomous exploration, while retaining its broad generalization ability?
Thus, a fundamental question arises: if human annotation limits the interaction intelligence of LLMs, can we instead encourage them to autonomously explore the environment with minimal supervision, thereby acquiring interaction capabilities and robust generalization?

Inspired by human cognition, we introduce \textbf{\framework}, a novel outcome-driven reinforcement learning framework that endows LLMs with interaction ability through autonomous exploration with minimal supervision. Unlike previous approaches that depend on dense reward shaping or large demonstration corpora, our framework integrates three key modules: (1) Group Rollout with In-Environment Interaction that enables efficient parallel exploration of environments, allowing the model to accumulate diverse interaction experiences; (2) a sparse reward structure based solely on task completion, preventing reward hacking while encouraging the development of genuine environmental understanding through interaction; and (3) the Interactive Policy Optimization~(IPO) algorithm that leverages group-normalized advantages to efficiently learn from grouped trajectories while maintaining reference to the original model.

Extensive experiments demonstrate \framework's effectiveness, achieving 97.78\% and 79.92\% cumulative reward on the ALFWorld and ScienceWorld benchmarks, significantly outperforming existing approaches. More remarkably, the model exhibits robust generalization, with only a -3.33\% generation gap in unseen environments—substantially lower than comparable methods. These results validate our framework's capability to foster genuine environmental understanding and adaptive behavior.

In summary, the main contributions of this work are as follows:

\begin{itemize}

    \item We introduce \framework, a framework that integrates outcome-driven reinforcement learning with LLMs for interactive planning, enabling robust interactive planning.
    \item We develop three technical modules: (1) a parallel group-based environmental interactive rollout mechanism for enhanced trajectory diversity, (2) a completion-driven reward architecture aimed at reducing reward hacking and encouraging autonomously exploration, and (3) Interactive Policy Optimization~(IPO) algorithm for effective policy optimization in multi-turn interaction.
    \item Through extensive evaluation, we demonstrate state-of-the-art performance on ALFWorld (97.78\%) and ScienceWorld (79.92\%) benchmarks, with strong generalization to unseen environments~(only -3.33\% generation gap) and significant reduction in invalid action generation.
    % \item We will release our implementation and datasets to support future research in interactive planning.
    
\end{itemize}

\section{Preliminaries}

Following established researches~\citep{qiao2024wkm}, we formalize embodied planning tasks within the framework of a Partially Observable Markov Decision Process (POMDP) due to the agent's inability to directly observe the environment's complete state. Within this framework, the agent interacts within a partially observable environment and develops plans based on environmental feedback. The POMDP is defined by a 7-tuple: 

\begin{equation}
    ( S, A, \Psi, \mathcal{M}, O, R, \gamma),
\end{equation}

where $S$ represents the state space defined by the environment, $A$ denotes the action space available to the agent, $O$ constitutes the observation space of the agent, $\Psi: S \times A \rightarrow S$ is the state transition function determining how actions will have effect on environmental states, $\mathcal{M}: S \rightarrow O$ is the observation function of the agent. Due to partial observability constraints, the agent cannot directly access states $s \in S$, but rather obtains partial observations of the current environment via the observation function, such that $o = \mathcal{M}(s)$, where $ o \in O$ and $s\in S$. Additionally, $R$ represents the reward function, and $\gamma$ is the discount factor.

In the context of planning tasks, the agent receives a task $q$, typically expressed as a natural language instruction. The agent's objective is to complete this instruction through environmental interaction. During this interaction process, the agent initially obtains an observation $o_0$ from the environment and initializes a trajectory $ \tau_0 = (q, o_0) $. At each time step $t$, the policy model $\pi_{\theta}$ generates the current action based on the historical trajectory: 

\begin{equation}
    a_t = \pi_{\theta}(\tau_{t}) = \pi_{\theta}(q,o_0,a_0,o_1,...,a_{t-1},o_{t}).
\end{equation}

%\jjg{
When performed, the action will lead to a change to the environment, resulting in a new state $s_{t+1} = \Psi(s_t, a_t)$, Following this transition, the agent receives a new partial observation $o_{t+1} = \mathcal{M}(s_{t+1})$. Subsequently, the agent's trajectory, is updated to $\tau_{t+1} = (q, o_0, a_0, o_1, ..., a_t, o_{t+1})$, incorporating all past actions and observations with the new observation. 
The reward for the agent is defined with a binary signal, where the agent receives a reward of 1 if the new state $s_{t+1}$ meets the predefined goal conditions specified by $q$, otherwise the reward is 0. 
%This sparse reward poses a significant challenge for the learning process because the provided minimal feedback requires the agent to develop more advanced exploration and planning strategies to learn efficiently from limited positive feedback.
This sparse reward, with minimal human priors, encourages the agent to autonomously explore the environment and develop advanced exploration and planning strategies during training.

%}

%The action will take effect in environment and result in a new state $_{t+1} = \Psi(s_t, a_t)$, subsequently, the agent will receive a new partial observation $o_{t+1} = \mathcal{M}(s_{t+1})$.
%After that, the trajectory will be updated to $\tau_{t+1} = (g,o_0,a_0,o_1,...,a_{t},o_{t+1})$. The reward is designed as a binary signal: a reward of 1 is assigned if the state $s_{t+1}$ satisfies the goal conditions specified by $g$; otherwise, the reward is 0. This sparse reward structure creates a challenging learning environment that necessitates sophisticated exploration and planning strategies.

% and $\textcolor{orange}{a_t} = \text{action\_extract}(m_t)$ is the action for $t$ step extracted from the thought context $m_t$.
% }

% The agent's response is structured into two distinct components, as illustrated in the orange font in Figure~\ref{fig:traj}: a thinking component that employs chain-of-thought processing to develop unrestricted reasoning and an action component that generates appropriate commands tailored to the action space of different environments based on specific tasks. 
% We provide detailed prompt designs for each task in Appendix~\ref{sec:prompt_design}. Importantly, we impose no restrictions on the generation space of the policy model, ensuring sufficient flexibility for exploration. Actions are extracted from the model's response by a dedicated action extraction function, which are then executed in the environment to obtain observations for the next time step.

\section{\framework}

\begin{figure}[t]
    \centering
    \includegraphics[width=\linewidth]{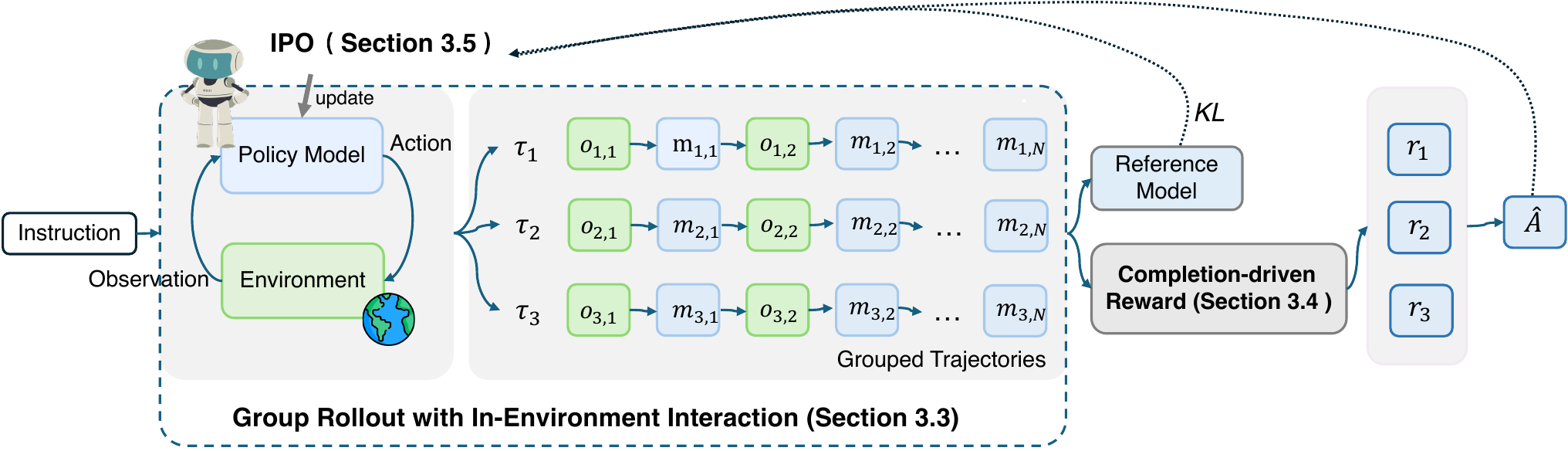}
    \caption{The \framework~consists of three tightly integrated core components: 1) Group Rollout with Environment Interaction, a parallel sampling strategy that simultaneously generates multiple trajectories to form trajectory groups; 2) Completion-driven Sparse Reward; 3) Interactive Policy Optimization algorithm that redesigns probability ratios and advantage allocation, and updates the policy function based on group relative rewards. The KL divergence between policy and reference model serves as a regularization term to prevent over-optimization.}
    \label{fig:main}
\end{figure}

\subsection{Overview}

In this section, we introduce \textbf{\framework}, an end-to-end training framework designed for long-term planning. \framework\ learns solely from outcome-based reward signals without requiring additional human supervision. Our key insight is to unleash the latent reasoning and planning capabilities of large language models (LLMs) through direct interaction with the environment, allowing them to self-evolve via reinforcement learning. To better model the interaction process, we formulate it as \textbf{ReAct-style trajectories} (Section~\ref{sec:react}), where progress is made through continuous `$Thought \rightarrow  Action \rightarrow  Observation$' cycles until task completion. \framework~ consists of three tightly integrated core components: Firstly, to address the problem of sparse rewards, we develop a parallel sampling strategy that simultaneously generates multiple trajectories forming trajectory groups to establish stable statistical baselines, which we call \textbf{Group Rollout with In-Environment Interaction}~(Section~\ref{sec:rollout}). Secondly, we design a pure \textbf{Completion-based Sparse Reward}~(Section~\ref{sec:reward}) structure that encourages development of genuine environmental understanding through autonomous interaction. Finally, and most importantly, we introduce the \textbf{Interactive Policy Optimization} (IPO) algorithm~(Section~\ref{sec:ipo}). Unlike Group Relative Policy Optimization (GRPO), which only handles single-turn dialogue, IPO is specifically designed for ReAct-style multi-turn interactions as its POMDP-oriented extension. While maintaining critic-free and group-relative baselines, we redesigned probability ratios and advantage allocation to handle multi-turn ReAct trajectories and extremely sparse rewards, effectively addressing probability degradation in long-sequence training and achieving efficient reward distribution.

\subsection{ReAct Paradigm}\label{sec:react}

\begin{wrapfigure}{r}{0.4\textwidth}
\centering
    \vspace{-2em}
    \includegraphics[width=0.4\textwidth]{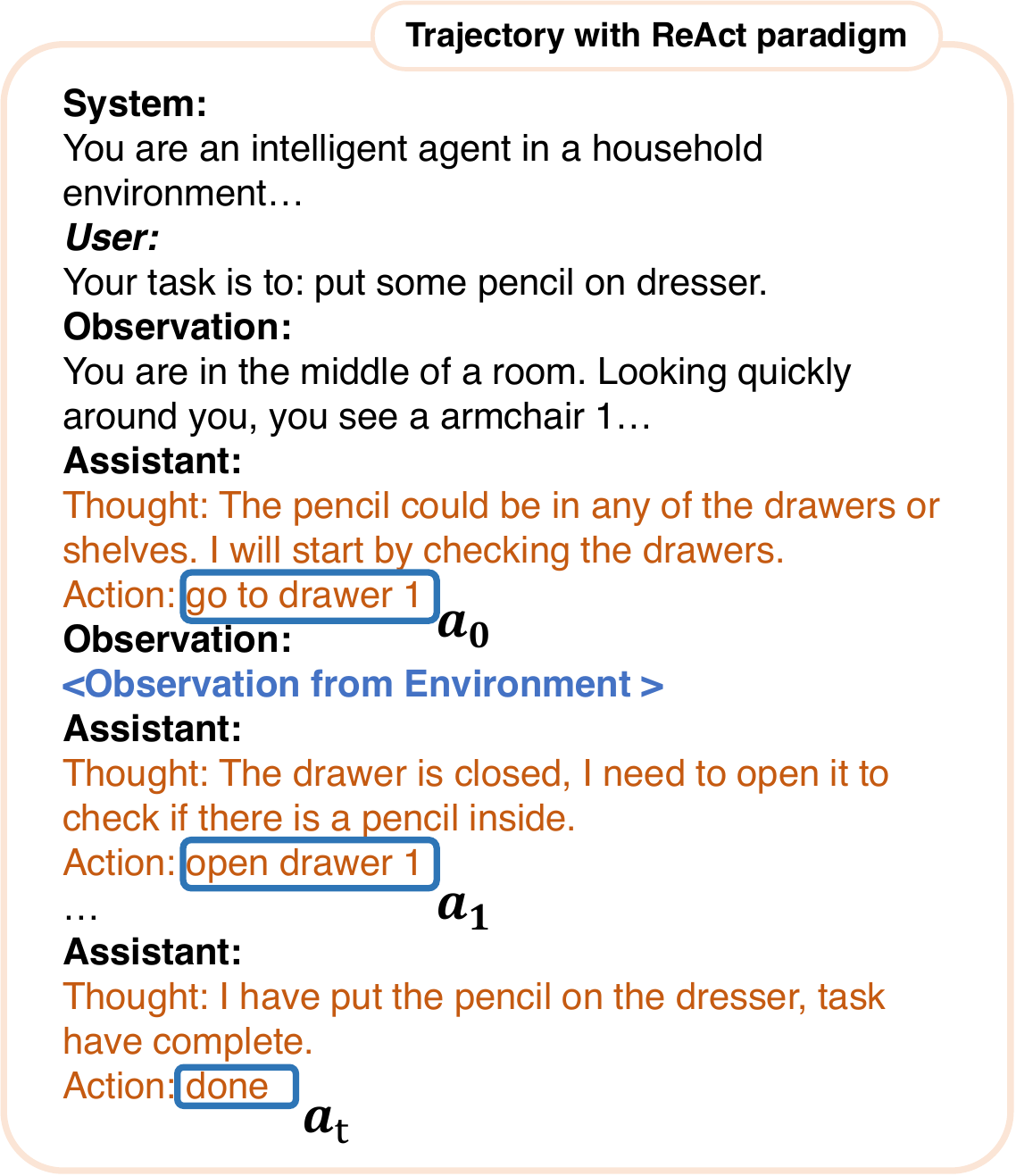}
    \caption{We employ the ReAct paradigm during generation and extract the action from the response $m_t$.}
    \label{fig:traj}
    \vspace{1em}
\end{wrapfigure}

As described above, at time step $t$, the agent generates an action based on a trajectory that includes environmental observations and previous actions. However, mapping directly from such a trajectory to action demands complex reasoning capabilities, like task goal decomposition, common sense knowledge application, and extraction of key information from observations is exceptionally challenging. Following \citet{yao2023react}, we adopt the ReAct paradigm that introduces the thought process \citep{wei2022cot} into LLM-based planning trajectories, as demonstrated in Figure~\ref{fig:traj}. Unlike traditional approaches, with the historical trajectory as input, the agent generates a thought context:

\begin{equation}
    \begin{aligned}
        \tau_{t+1} = (q, o_0, \textcolor{orange}{m_0}, o_1, ..., \textcolor{orange}{m_t}, o_{t+1}),
    \end{aligned}
\end{equation}

Where $\textcolor{orange}{m_t}$ denotes the model's response for $t$-th step, it consists of thoughts and a text-form action:

\begin{equation}
    \begin{aligned}
        m_t = (\phi_t, a_t).
    \end{aligned}
\end{equation}

We extract action $a_t$ from $m_t$, which is then used to interact with the environment to obtain the new observation $o_{t+1}$.

\subsection{Group Rollout with In-Environment Interaction}\label{sec:rollout}

To effectively train an LLM for solving interactive environment tasks, we introduce a parallel-rollout procedure that enhances exploration efficiency. We begin by sampling $K$ task-environment pairs from distribution $\mathcal{D}$ and creating $n$ parallel replicas for each pair. Each replica is initialized with its reset environment and an empty trajectory buffer. For up to $\textit{max\_steps}$ iterations, the frozen policy $ \pi_{\theta_{\text{old}}} $ generates responses conditioned on both the task prompt and current trajectory. From each response, we parse an action. To enhance the model's planning capabilities and effectively prevent the generation of excessive invalid actions, we introduce a special ``done'' action to terminate the replica. The model can use ``done'' when it determines that the task is either unsolvable or already completed; otherwise, the action executes in the environment, generating a new observation that gets appended to the trajectory along with the response. Once all replicas terminate or reach the step limit, we aggregate all trajectories into the global buffer $\mathcal{T}$, compute returns, and update the policy according to Eqs.(\ref{equal:ipo})--(\ref{eq:advantage}). We provide the algorithm pseudo code in Appendix~\ref{sec:rollout_code}. This approach efficiently collects diverse multi-turn ReAct trajectories, establishing a robust foundation for policy optimization that generalizes effectively to complex long-horizon planning challenges.

\subsection{Completion-driven Reward}\label{sec:reward}

% To teach the model how to interact with environment, we use a rule-based completion reward to optimize the model:
To encourage autonomous exploration, we use only a rule-based completion reward without further restriction to optimize the model:

\begin{equation}
r =
    \begin{cases}
    0 & \text{if the task is incomplete,} \\
    1 & \text{if the task is completed}.
    \end{cases}
\end{equation}

Unlike previous approaches using GRPO~\citep{guo2024deepseekr1} that introduced format rewards or length penalty, we do not explicitly provide such constraints. 
% On the one hand, we found that LLMs can maintain a clear response structure with guidance from the system prompt. On the other hand, in the planning task, the LLM must output the proper format~(like `Action:' in Figure~\ref{fig:traj}) and generate the action which is included in the specific action space, thereby interacting with the environment and completing the task. 
On the one hand, we found that LLMs can maintain a clear response structure with guidance from the system prompt, and only a parsable and valid response can interact with the environment to complete the task. On the other hand, we argue that reducing human priors in reward design can lower the risk of reward hacking and encourage the model to explore autonomously, generating more flexible and diverse trajectories.
% Therefore, we believe that Outcome-supervised Reward can implicitly guide the LLMs to form appropriate output structures, achieving more natural and effective environmental interaction capabilities. 
Therefore, we believe that outcome-based reward can implicitly guide LLMs to develop appropriate output structures, while allowing greater flexibility and freedom during exploration to encourage more natural and effective interactions with the environment.

\subsection{Interactive Policy Optimization (IPO)}\label{sec:ipo}

To enable efficient environment interaction and planning while substantially reducing training cost, we introduce Interactive Policy Optimization~(IPO). IPO relies on the group rollout with In-Environment Interaction to make the policy model $\pi_{\theta}$ generate multiple trajectories. We employ the group-normalized advantage estimator, which has no learnable critic model, but its probability ratio is calculated over the designated trajectories tokens prefix up to step $t$, including all preceding thought $\phi$ and action $a$ tokens:

\begin{equation}
    Pr_{t}(\theta) = \frac{\pi_\theta\left(\phi_t, a_t \mid \tau_{i,<t}\right)}{\pi_{\theta_{\text {old}}}\left(\phi_t, a_t \mid \tau_{i,<t}\right)}.
\end{equation}

We optimize the policy model by maximizing the following objective function:

\begin{equation}
    \begin{aligned}
        & \mathcal{J}_{\text{IPO}}(\theta)=\mathbb{E}_{(q,e) \sim \mathcal{D},\left\{\tau_i\right\}_{i=1}^n \sim \pi_{\theta_{old}}(\mathcal{T} \mid (q,e))} \\
        & \frac{1}{n} \sum_{i=1}^n \frac{1}{\left|\tau_i\right|} \sum_{t=1}^{\left|\tau_i\right|}\left\{\min \left[Pr_{t}(\theta) \hat{A}_{i, t}, \operatorname{clip}\left(Pr_{t}(\theta), 1-\varepsilon, 1+\varepsilon\right) \hat{A}_{i, t}\right]-\beta \mathbb{D}_{K L}\left[\pi_\theta| | \pi_{\text{ref}}\right]\right\},
    \end{aligned}\label{equal:ipo}
\end{equation}

where $\hat{A}_{i, t}$ is the advantage of $i$-th trajectory in the group for time step $t$, which is calculated as follows:

\begin{equation}
\label{eq:advantage}
    \hat{A}_{i,t} = \frac{r_i - \mu_r}{\sigma_r}.
\end{equation}

Where $t$ represents the $t$-th step of the trajectories, $\mu_r$ is the mean reward of the group, and $\sigma_r$ is the reward deviation of the whole group. Additionally, the objective of IPO integrates a KL divergence penalty term between the training policy and the reference policy to prevent the model from deviating too much~(Eq.\ref{equal:ipo}).

\begin{table}[t]
  \centering
  \vspace{-1em}
  \caption{\textbf{Main Results.} The best results are marked in \textbf{bold}.\dag ~indicates vanilla models tested with ReAcT prompting, and * denotes the best results reported in the original papers.  }\label{tab:main_results}
  \resizebox{\textwidth}{!}{
    \begin{tabular}{lcccccc}
    \toprule
    \multirow{2}[2]{*}{Method} & \multicolumn{3}{c}{ALFWorld} & \multicolumn{3}{c}{ScienceWorld} \\
           & Seen  & Unseen & Avg. & Seen  & Unseen & Avg. \\
    \midrule
    \normalrow
    \multicolumn{7}{l}{\textit{Referenced Results (Not Directly Comparable)}} \\
    \midrule
    CoLA* \citep{Jia2025ControllingLL} & 77.9 & 74.6 & 76.3 & 28.4 & 21.8 & 25.1 \\
    IPR* \citep{xiong-etal-2024-ipr} & 70.3 & 74.7 &  72.5 & - & - & - \\
    STeCa*  \citep{Wang2025STeCaST} & 74.3 & 76.1 &  75.2 & - & - & - \\
    OREO* \citep{wang2024oreo} & 79.1 & 80.7 & 79.9 & - & - & - \\
    QLASS* \citep{Lin2025QLASSBL} & 77.9 &  82.8 & 80.4 & 75.3 & 66.4 & 70.9 \\
    KnowAgent* \citep{zhu-etal-2025-knowagent} & 66.71 & 62.69 & 64.70 & 58.67 & 49.18 & 53.93 \\
    WKM* \citep{qiao2024wkm}  & 68.57 & 65.93 & 67.25 & 58.67 & 49.18 & 53.93 \\
    \midrule
    \normalrow
    \multicolumn{7}{l}{\textit{Our Implementation}} \\
    \midrule
    GPT-3.5-Turbo\dag \citep{tom20gpt3} & 7.14  & 7.46  & 7.30 & 28.03 & 21.63 & 24.83 \\
    GPT-4o\dag \citep{openai2023gpt4} & 58.57 & 52.24 & 55.41 & 59.6  & 56.75 & 58.18 \\
    Deepseek-v3\dag \citep{deepseek2024deepseekv3} & 36.43 & 31.34 & 33.89 & 29.28 & 27.45 & 28.37 \\
    InternLM 2.5 20B Chat\dag \citep{cai24internlm2} & 7.86  & 11.94 & 9.90 & 33.66 & 31.15 & 32.41 \\
    Qwen2.5-7B-Instruct\dag \citep{yang2024qwen2_5} & 30.00 & 32.09 & 31.05 & 28.01 & 16.09 & 22.05 \\
    Qwen2.5-7B-Instruct + SFT\citep{zeng2024agenttuning}   & 80.71 & 79.10  & 79.91 & 68.81 & 55.7  & 62.26 \\
    Qwen2.5-7B-Instruct + NAT\citep{wang2024nat}  & 63.57 & 66.42 & 65.00 & 58.56 & 49.75 & 54.16 \\
    Qwen2.5-7B-Instruct + ETO\citep{song2024eto} & 84.29 & 79.85 & 82.07 & 55.08 & 57.32 & 56.20 \\
    \rowhighlight
    Qwen2.5-7B-Instruct + \framework(Ours) & \textbf{99.29} & \textbf{96.27} & \textbf{97.78} & \textbf{83.66} & \textbf{76.18} & \textbf{79.92} \\
    \bottomrule
    \end{tabular}
    }
\end{table}

\section{Experiments}

\subsection{Implementation}

\paragraph{Evaluation} To assess model capabilities in interactive planning tasks, we conducted systematic evaluations using two text-based embodied world simulator benchmarks: ALFWorld~\citep{mohit2021alfworld} and ScienceWorld~\citep{wang2022scienceworld}. ALFWorld encompasses six categories of planning tasks set primarily in home environments, covering not only basic object manipulation (such as ``pick and place'') but also tasks requiring complex interaction sequences. ScienceWorld presents a more challenging benchmark, requiring models to complete scientific experiments in a highly interactive environment. Each task was evaluated only once, with interaction steps limited to 30 to ensure evaluation efficiency and consistency. More evaluation and benchmark details will introduced Appendix \ref{app: benchamrk}.

\paragraph{Baseline} To benchmark our \framework\ against other training-based approaches, we compared it with three state-of-the-art methods that incorporate human priors: SFT~\citep{zeng2024agenttuning}, which fine-tunes on expert trajectories to learn interaction capabilities; NAT~\citep{wang2024nat}, which incorporates learning from rejected trajectories; and ETO~\citep{song2024eto}, which learns from expert trajectories through DPO. Our evaluation also includes larger models: GPT-3.5 Turbo~\citep{tom20gpt3}(gpt-3.5-turbo-0125), GPT-4~\citep{openai2023gpt4}(gpt-4o-2024-08-06), DeepSeek-V3~\citep{deepseek2024deepseekv3}(DeepSeek-V3-0324), and InternLM 2.5 20B~\citep{cai24internlm2}, to demonstrate the current performance of foundation large language models (both open-source and closed-source) on interactive planning tasks. All our implemented training-based methods are exclusively fine-tuned on Qwen2.5-7B-Instruct~\citep{yang2024qwen2_5}, and all models are evaluated using a consistent set of prompts. Additionally, while many methods cannot be directly compared due to differences in models and evaluation approaches, we still provide these results in Table~\ref{tab:appendix_results} for reference.

\paragraph{Training Details}

We trained our model based on verl\footnote{\url{https://github.com/volcengine/verl}}, a flexible LLM reinforcement learning training library, and implemented an interaction module on this foundation. In each training step, we sampled 128 tasks, with 5 trajectories sampled for each task, and each trajectory containing up to 30 interaction steps. All experiments were conducted on 8 NVIDIA A100 80GB GPUs. Additionally, we used Qwen2.5-7B-Instruct~\citep{yang2024qwen2_5}\footnote{\url{https://huggingface.co/Qwen/Qwen2.5-7B-Instruct}} as the backbone model, with training performed at a context length of 4096. 

\subsection{Main results}

As shown in Table~\ref{tab:main_results}, we conducted comprehensive evaluations of prompt-based and training-based approaches on two challenging interactive environments, ALFWorld and ScienceWorld. The experimental results reveal several key findings:

First, current state-of-the-art foundational LLMs demonstrate notable limitations in interactive planning capabilities. While GPT-4 achieved the best performance~(55.41\% on ALFWorld and 58.18\% on ScienceWorld), it still lags behind training-based approaches. GPT-3.5-Turbo showed significantly lower performance (7.30\% and 24.83\% respectively), indicating the challenges large language models face in environmental interaction tasks.

Compares with other baseline, \framework\ demonstrates consistent improvements across all evaluation metrics. On the ALFWorld benchmark, it achieved accuracy rates of 99.29\% and 96.27\% in seen and unseen scenarios, respectively, surpassing the previous best method ETO by 15.00\% and 16.42\%. On the ScienceWorld dataset, \framework\ achieved accuracy rates of 83.66\% and 76.18\% in seen and unseen scenarios, outperforming competitive methods by 28.58\% and 18.86\%, respectively. The performance improvement from the Qwen2.5-7B-Instruct (averaging 31.05\% on ALFWorld and 22.05\% on ScienceWorld) to \framework\ demonstrates the effectiveness of our reinforcement learning framework. 

We evaluate three trajectory-based baselines in ALFWorld that leverage human prior knowledge to construct training datasets and observe their performance clustering around 80\%, suggesting a performance ceiling imposed by the reliance on human-annotated data. In contrast, \framework\ easily surpasses this ceiling, achieving over 95\% by leveraging reinforcement learning, which enables the model to explore the environment autonomously and generate more diverse and flexible data.

Overall, these findings highlight several important issues: foundational LLMs exhibit significant deficiencies in interactive planning tasks, and while trajectory-based offline training helps narrow this gap, it fails to completely resolve it. \framework\ introduces a novel paradigm based on online exploration, enabling models to achieve exceptional interactive planning capabilities through autonomous exploration. This approach establishes a robust new baseline for future research in interactive planning.

\subsection{Analysis}

\begin{figure}[t]
    \centering
    \includegraphics[width=0.99\linewidth]{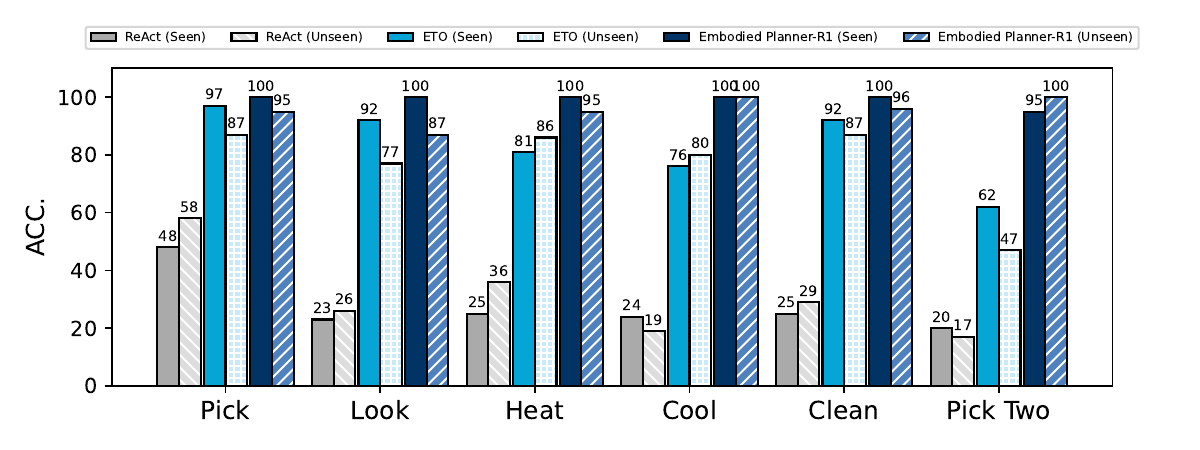}
    \caption{Completion rates of Qwen2.5-7B-Instruct on various ALFWorld tasks, covering both seen and unseen scenarios. We compare the prompt-based method ReAct with the training-based method ETO and \framework.}\label{fig:alfworld_performance}
\end{figure}

\paragraph{\framework\ maintains high-level performance even on more difficult problems and unseen environments.} Figure~\ref{fig:alfworld_performance} compares the performance differences among various methods across different ALFWorld tasks, revealing clear difficulty hierarchies and capability differentials among the methods. As shown in the Figure, the ``Pick'' single-object task is relatively the most basic, where all training-based methods perform well, while purely prompt-based methods achieve less than 60\% task completion rate, indicating that current LLMs still have optimization potential even for simple interactive planning tasks. The ``Look'', ``Cool'', ``Clean'', and ``Heat'' tasks present moderate difficulty as they involve more complex state transitions. The ``Pick Two'' task is clearly the most challenging, with ReAct performing worst on this task (only 20.8\% in seen and 17.6\% in unseen), and ETO also showing significant decline (62.5\% in seen, 47.05\% in unseen). Only \framework\ maintains high-level performance (95.8\% in seen, 100\% in unseen). This difficulty distribution reflects the challenges drastically increase when tasks involve more object interactions, complex state transitions, or environmental reasoning. However, \framework\ maintains stable performance across the different difficulty levels, demonstrating that our method enables models to learn how to handle complex long-horizon planning tasks.

\paragraph{Generalization capabilities in unseen environments.} 

To evaluate the generalization capability of different methods, we define the generalization gap $\Delta(abs)$ as the difference between task completion rates on unseen and seen tasks: $\Delta(abs) = Acc_{unseen} - Acc_{seen}$, as shown in Table~\ref{tab:generation_gap}. Experimental results demonstrate that our method exhibits significant advantages in generalization performance. Overall, our approach achieves a smaller average generalization gap~(-3.33\%) compared to the baseline method ETO~(-7.34\%). Notably, in the complex ``Pick Two'' task, our method not only overcomes the generalization challenge but achieves positive improvement~(+4.38\%), while ETO shows substantial performance degradation~(-24.72\%). This superior generalization ability is particularly evident in handling complex tasks. Further analysis reveals that our method also demonstrates stronger stability, with a relatively smaller performance variance across different task types~(ranging from -12.50\% to +4.38\%). Although the ``Look'' task poses considerable challenges for both methods, our approach maintains a relatively smaller performance drop~(-12.5\%). These results indicate that \framework, by encouraging the model to explore the environment autonomously, effectively enhances the model's adaptability to variations in task scenarios while maintaining cross-task stability.

\begin{table}[t]
  \centering
  \caption{ETO and \framework's relative improvement on unseen tasks compared with seen tasks in ALFWorld, where \framework\ maintains strong generalization ability even on unseen tasks.}\label{tab:generation_gap}%
    \begin{tabular}{cccccccc}
    \toprule
          & Pick  & Look  & Heat  & Clean & Cool  & Pick Two & Avg. \\
    \midrule
     ETO  & -9.89 & -15.82 & +6.95  & -5.94 & +5.39  & -24.72 & -7.34 \\
    Ours  & -4.20 & -12.50 & -4.55 & -3.13 & 0.00  & +4.38  & -3.33 \\
    \bottomrule
    \end{tabular}%
\end{table}%

\paragraph{Efficient planning through invalid action reduction.}

\begin{figure}[t]
    \centering
    
    % 三图并排，每个占用总宽度的30%，留有适当间隔
    \begin{minipage}[t]{0.64\textwidth}
        \begin{subfigure}[b]{0.49\linewidth}
        \centering
        \includegraphics[width=\linewidth]{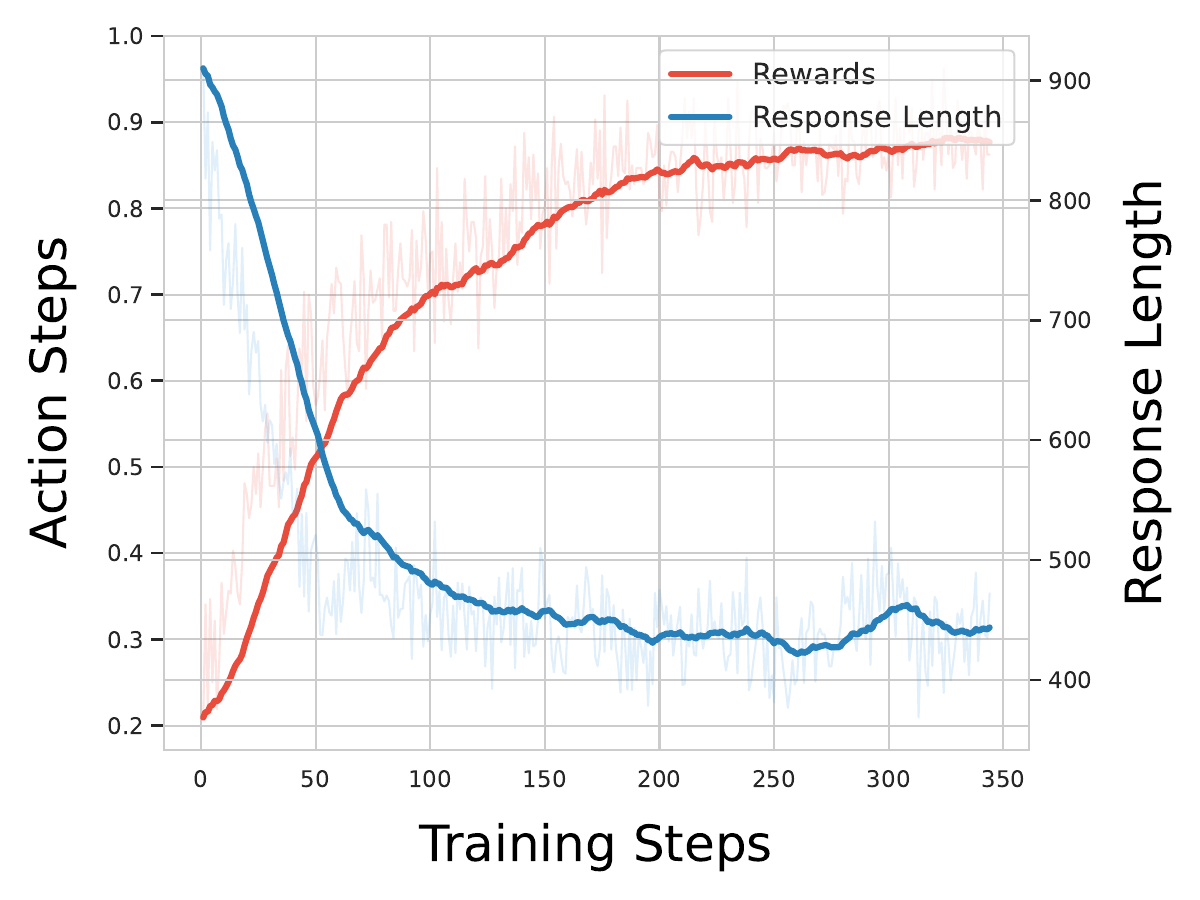}
        \caption{Rewards and response length}
        \label{fig:sciworld_progress}
    \end{subfigure}
    \begin{subfigure}[b]{0.49\linewidth}
        \centering
        \includegraphics[width=\linewidth]{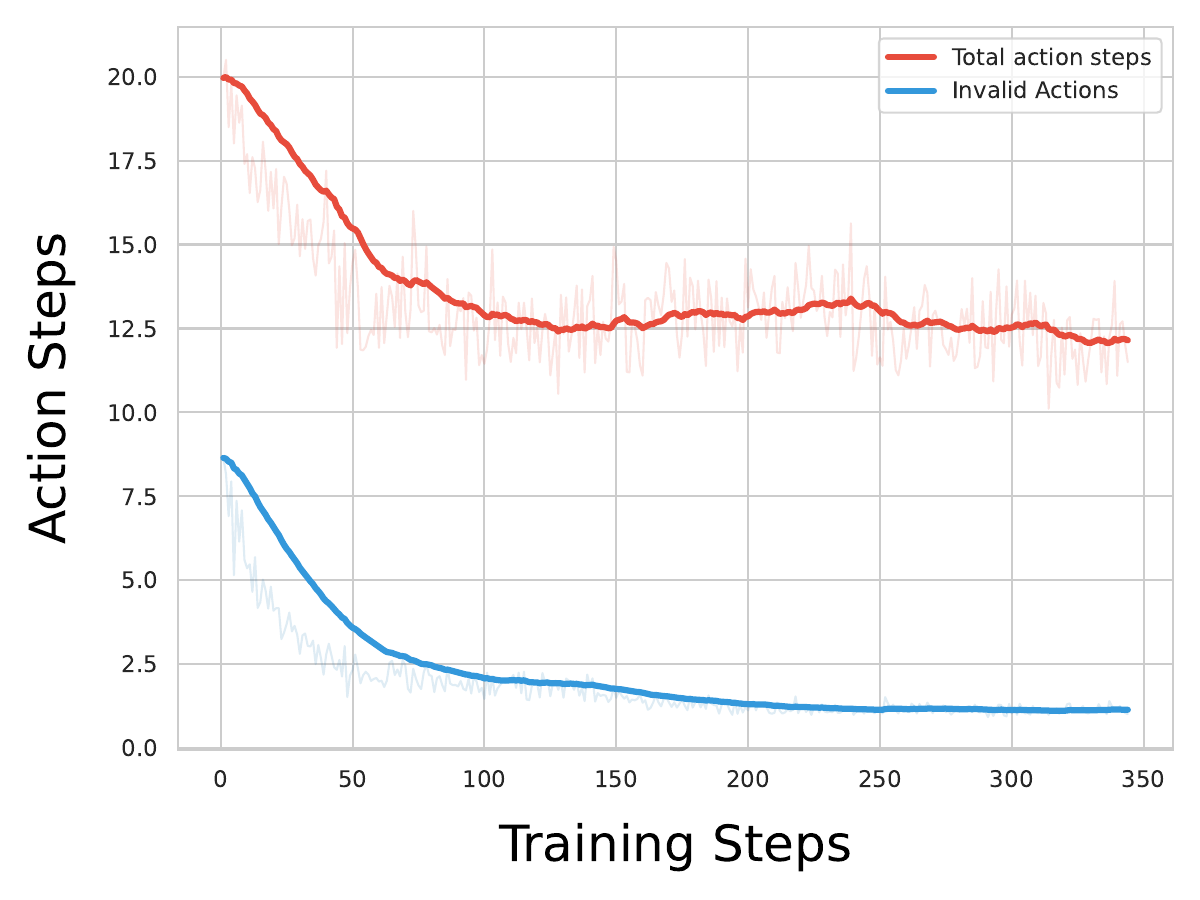}
        \caption{Total and invalid action steps}
        \label{fig:sciworld_steps}
    \end{subfigure}
    \caption{Training progress of \framework\ in ScienceWorld: (a) shows the rewards and total response length in multi-round exploration; (b) illustrates the decreasing trends in both total and invalid action steps during training.}\label{fig:sciworld_combined}
    \end{minipage}
    \hfill  % 自动水平间距
    \begin{minipage}[t]{0.35\textwidth}
        \centering
        \includegraphics[width=\textwidth]{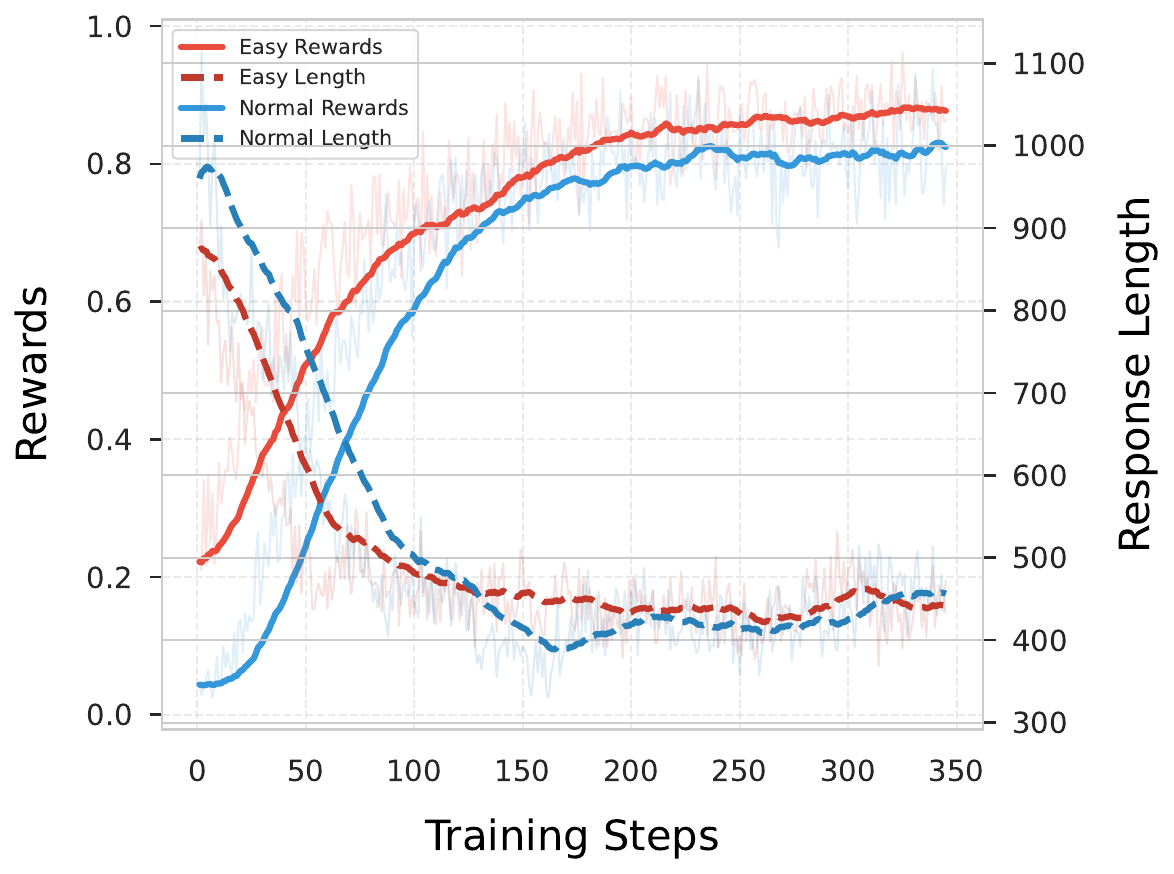}
        \caption{Comparison of rewards and response lengths between ``Easy'' and ``Normal'' mode in ScienceWorld.}
        \label{fig:diffculty}
    \end{minipage}
    \vspace{-2em}
%     \caption{Training progress of Embodied-R1 in ScienceWorld: (a) shows the rewards (task completion
% rate) and total response length in multi-round exploration; (b) illustrates the decreasing trends in both
% total and invalid action steps during training. (c) demonstrates similar learning performance across different difficulty levels}
%     \label{fig:sciworld_combined}
\end{figure}

Figure~\ref{fig:sciworld_combined} demonstrates how \framework\ achieves efficient planning by systematically reducing invalid actions during training in the ScienceWorld. The reward curve shows steady improvement as the model learns to generate more valid action sequences~(see Figure~\ref{fig:sciworld_progress}). Notably, the average response length decreases from approximately 900 tokens to 500 tokens by step 50, reflecting more concise and efficient action planning. This efficiency gain is directly linked to the model's ability to avoid invalid actions, as shown in Figure~\ref{fig:sciworld_steps}. The total number of action steps~(blue line) reduces from an initial 20 steps to a consistent 12.5 steps, while invalid actions~(yellow line) are nearly eliminated after 50 training steps. This substantial reduction in invalid actions indicates that \framework\ has developed a robust understanding of environmental constraints, leading to more efficient planning strategies. Instead of generating actions that would be rejected by the environment, The model learns to propose valid action sequences and formulate more efficient plans, thereby reducing the total steps needed for task completion. These results demonstrate that \framework\ achieves efficient planning primarily through its ability to avoid invalid actions and systematic exploration, leading to more streamlined and effective task execution.

\paragraph{Increase difficulty in environment.}

Figure~\ref{fig:diffculty} compares the learning performance of agents in ScienceWorld under two difficulty settings~(``Easy'' and ``Normal''). The ``Easy'' mode uses all five simplification features to reduce low-level operational demands, which were used in the main experiment and previous methods. In contrast, the ``Normal'' mode preserves the complete mechanics, requiring agents to navigate, open containers, water plants, and execute all fundamental actions, which poses higher demands on the model. As shown in Figure~\ref{fig:diffculty}, the LLM demonstrates notable learning progress in both settings. Under the ``Easy'' setting, the LLM exhibits a faster learning progress, reaching a reward value of 0.6 at approximately 50 steps, whereas the ``Normal'' environment requires nearly 100 steps to achieve comparable performance. In the later training stages, the ``Easy'' setting yields higher stable reward (approximately 0.88), compared to only about 0.82 in the ``Normal'' setting. Concurrently, the model's response length shows a similar downward trend. While initially the ``Normal'' response length was higher (approximately 1100), they eventually both converge to around 450. The semi-transparent regions surrounding the curves indicate that both settings display higher variance during early training phases, gradually stabilizing as training progresses, though the ``Normal'' environment consistently maintains greater fluctuations. 

These results further demonstrate that by encouraging the model to autonomously explore environment, \framework\ enables the model to adopt more flexible and diverse exploration strategies. This allows it to adapt even in more challenging settings, where it learns to master complex actions through exploration and develops enhanced planning capabilities.

\section{Related Work}
\label{app: relatedwork}
\subsection{Long-Horizon planning with large language models}
The emergence of LLMs has significantly advanced long-horizon planning~\citep{liu2024embodiedsurvey,xi2025agentsurvey,duan2022embodiedsurvey,guo2024agentsurvey,zhou2024agentsurvey}, enabling models to explore interactive environments and accomplish complex goals autonomously. Current methods can be categorized into two primary approaches: prompt-based methods and training-based methods. Prompt-based methods utilize manually crafted instructions or exemplars to enhance LLMs' planning capabilities without additional training.~\citet{yao2023react} proposed ReAct, which guides models to interact with environments through a structured ``thinking-action'' paradigm, introducing explicit reasoning processes to improve environmental understanding and task performance. Subsequent research has expanded these foundations by developing specialized skill sets~\citep{Nottingham2024sso, qiao2024wkm, zhao2023ck,yao2024retroformer} and optimizing prompt engineering techniques~\citep{shinn2023reflexion, Sarch2024ICAL, sun2023adaplanner}. While these approaches benefit from flexible deployment without retraining, they demonstrate limited effectiveness in highly complex environments.
Training-based methods directly optimize LLMs for environmental interaction through additional training. A common approach is to use behavioral cloning, where an agent learns a policy by mimicking expert trajectories~\citep{yin2024lumos, zeng2024agenttuning,chen2024fireact,qiao2024wkm,yuan2025agentr,song2024agentbank,xi2024agentgym}. To obtain more diverse data for model training, some research has also explored incorporating negative trajectories into the training process~\citep{song2024eto,wang2024nat,putta2024agentq, wang2025d2po, xiong2024wes,zhai2025enhancing}. However, these methods predominantly rely on pre-collected static datasets, which introduce critical limitations: (1) insufficient diversity in training examples, and (2) distribution shifts between training and test environments.
Our proposed \framework\ addresses these limitations by generating training data through online environmental interaction. This dynamic approach allows models to actively explore environments and continuously optimize strategies based on direct feedback.

\subsection{Self-Evolution via Reinforcement Learning}

Reinforcement Learning has been proven to be an effective methodology for enabling LLMs to achieve self-evolution across various capabilities~\citep{sun2023reasoning, li2025reasoningsurvey, lyu2025oreal, ying2024internlm-math, lightman2024lets, luo2023wizardmath}. Recently, DeepSeek-R1~\citep{guo2024deepseekr1} discovered that employing simple reward functions for reinforcement learning can effectively drive self-improvement in LLMs, achieving remarkable performance enhancements across various complex reasoning tasks. Building upon this foundation, numerous studies have adopted this approach to attain SOTA performance across multiple domains, including mathematics~\citep{hu2025open,lyu2025oreal, gao2025kimi}, logical reasoning~\citep{xie2025logic}, and even visual domains~\citep{liu2025segzero, pan2025medvlmr1}.

While these approaches demonstrate significant self-evolution in reasoning and adaptation within static contexts, interactive planning presents unique challenges that go beyond both single-turn reasoning and retrieval-based dialogues~\citep{zheng2025deepresearcher, jin2025search,li2025searcho1}, as they require continuous interaction with environments to update states and involve planning across diverse action spaces. In these dynamic environments, limited understanding and weak planning capabilities can cause models to get stuck in costly blind trials. Our proposed \framework\ extends self-evolution to these complex planning scenarios, enabling LLMs to iteratively refine their strategies through reinforcement learning as they explore uncertain environments with incomplete information.

\section{Conclusion}

In this paper, we propose \framework, a reinforcement learning framework that enables LLMs to achieve self-evolution in multi-turn environmental interaction planning tasks. Our approach demonstrates that models can autonomously develop effective planning strategies through carefully designed prompt constraints and outcome-based reward mechanisms, without relying on expert trajectories or extensive human priors. Experimental results across multiple environments show that \framework\ significantly outperforms existing baselines, validating the effectiveness of self-evolution via reinforcement learning in complex interactive scenarios. Notably, our analysis reveals that unlike single-turn reasoning tasks, multi-turn interactive planning benefits from shorter context lengths, as performance improves when ineffective actions are reduced. This suggests that interactive planning tasks prioritize efficient action planning over lengthy deliberation or redundant action planning. These findings may provide some insights for LLM-based interactive planning systems

\newpage

\bibliographystyle{plainnat}

\bibliography{reference}

\begin{thebibliography}{62}
\providecommand{\natexlab}[1]{#1}
\providecommand{\url}[1]{\texttt{#1}}
\expandafter\ifx\csname urlstyle\endcsname\relax
  \providecommand{\doi}[1]{doi: #1}\else
  \providecommand{\doi}{doi: \begingroup \urlstyle{rm}\Url}\fi

\bibitem[Brown et~al.(2020)Brown, Mann, Ryder, Subbiah, Kaplan, Dhariwal, Neelakantan, Shyam, Sastry, Askell, Agarwal, Herbert{-}Voss, Krueger, Henighan, Child, Ramesh, Ziegler, Wu, Winter, Hesse, Chen, Sigler, Litwin, Gray, Chess, Clark, Berner, McCandlish, Radford, Sutskever, and Amodei]{tom20gpt3}
Tom~B. Brown, Benjamin Mann, Nick Ryder, Melanie Subbiah, Jared Kaplan, Prafulla Dhariwal, Arvind Neelakantan, Pranav Shyam, Girish Sastry, Amanda Askell, Sandhini Agarwal, Ariel Herbert{-}Voss, Gretchen Krueger, Tom Henighan, Rewon Child, Aditya Ramesh, Daniel~M. Ziegler, Jeffrey Wu, Clemens Winter, Christopher Hesse, Mark Chen, Eric Sigler, Mateusz Litwin, Scott Gray, Benjamin Chess, Jack Clark, Christopher Berner, Sam McCandlish, Alec Radford, Ilya Sutskever, and Dario Amodei.
\newblock Language models are few-shot learners.
\newblock In Hugo Larochelle, Marc'Aurelio Ranzato, Raia Hadsell, Maria{-}Florina Balcan, and Hsuan{-}Tien Lin, editors, \emph{Advances in Neural Information Processing Systems 33: Annual Conference on Neural Information Processing Systems 2020, NeurIPS 2020, December 6-12, 2020, virtual}, 2020.
\newblock URL \url{https://proceedings.neurips.cc/paper/2020/hash/1457c0d6bfcb4967418bfb8ac142f64a-Abstract.html}.

\bibitem[Chen et~al.(2023{\natexlab{a}})Chen, Shu, Shareghi, Collier, Narasimhan, and Yao]{chen2024fireact}
Baian Chen, Chang Shu, Ehsan Shareghi, Nigel Collier, Karthik Narasimhan, and Shunyu Yao.
\newblock Fireact: Toward language agent fine-tuning.
\newblock \emph{CoRR}, abs/2310.05915, 2023{\natexlab{a}}.
\newblock \doi{10.48550/ARXIV.2310.05915}.
\newblock URL \url{https://doi.org/10.48550/arXiv.2310.05915}.

\bibitem[Chen et~al.(2023{\natexlab{b}})Chen, Cui, Chen, Tan, Zhang, Zhao, and Wang]{chen2023robogpt}
Yaran Chen, Wenbo Cui, Yuanwen Chen, Mining Tan, Xinyao Zhang, Dongbin Zhao, and He~Wang.
\newblock Robogpt: an intelligent agent of making embodied long-term decisions for daily instruction tasks.
\newblock \emph{CoRR}, abs/2311.15649, 2023{\natexlab{b}}.
\newblock \doi{10.48550/ARXIV.2311.15649}.
\newblock URL \url{https://doi.org/10.48550/arXiv.2311.15649}.

\bibitem[Choudhury(2025)]{choud2025agentprm}
Sanjiban Choudhury.
\newblock Process reward models for {LLM} agents: Practical framework and directions.
\newblock \emph{CoRR}, abs/2502.10325, 2025.
\newblock \doi{10.48550/ARXIV.2502.10325}.
\newblock URL \url{https://doi.org/10.48550/arXiv.2502.10325}.

\bibitem[DeepSeek{-}AI et~al.(2024)DeepSeek{-}AI, Liu, Feng, Xue, Wang, Wu, Lu, Zhao, Deng, Zhang, Ruan, Dai, Guo, Yang, Chen, Ji, Li, Lin, Dai, Luo, Hao, Chen, Li, Zhang, Bao, Xu, Wang, Zhang, Ding, Xin, Gao, Li, Qu, Cai, Liang, Guo, Ni, Li, Wang, Chen, Chen, Yuan, Qiu, Li, Song, Dong, Hu, Gao, Guan, Huang, Yu, Wang, Zhang, Xu, Xia, Zhao, Wang, Zhang, Li, Wang, Zhang, Zhang, Tang, Li, Tian, Huang, Wang, Zhang, Wang, Zhu, Chen, Du, Chen, Jin, Ge, Zhang, Pan, Wang, Xu, Zhang, Chen, Li, Lu, Zhou, Chen, Wu, Ye, Ye, Ma, Wang, Zhou, Yu, Zhou, Pan, Wang, Yun, Pei, Sun, Xiao, and Zeng]{deepseek2024deepseekv3}
DeepSeek{-}AI, Aixin Liu, Bei Feng, Bing Xue, Bingxuan Wang, Bochao Wu, Chengda Lu, Chenggang Zhao, Chengqi Deng, Chenyu Zhang, Chong Ruan, Damai Dai, Daya Guo, Dejian Yang, Deli Chen, Dongjie Ji, Erhang Li, Fangyun Lin, Fucong Dai, Fuli Luo, Guangbo Hao, Guanting Chen, Guowei Li, H.~Zhang, Han Bao, Hanwei Xu, Haocheng Wang, Haowei Zhang, Honghui Ding, Huajian Xin, Huazuo Gao, Hui Li, Hui Qu, J.~L. Cai, Jian Liang, Jianzhong Guo, Jiaqi Ni, Jiashi Li, Jiawei Wang, Jin Chen, Jingchang Chen, Jingyang Yuan, Junjie Qiu, Junlong Li, Junxiao Song, Kai Dong, Kai Hu, Kaige Gao, Kang Guan, Kexin Huang, Kuai Yu, Lean Wang, Lecong Zhang, Lei Xu, Leyi Xia, Liang Zhao, Litong Wang, Liyue Zhang, Meng Li, Miaojun Wang, Mingchuan Zhang, Minghua Zhang, Minghui Tang, Mingming Li, Ning Tian, Panpan Huang, Peiyi Wang, Peng Zhang, Qiancheng Wang, Qihao Zhu, Qinyu Chen, Qiushi Du, R.~J. Chen, R.~L. Jin, Ruiqi Ge, Ruisong Zhang, Ruizhe Pan, Runji Wang, Runxin Xu, Ruoyu Zhang, Ruyi Chen, S.~S. Li, Shanghao Lu, Shangyan Zhou,
  Shanhuang Chen, Shaoqing Wu, Shengfeng Ye, Shengfeng Ye, Shirong Ma, Shiyu Wang, Shuang Zhou, Shuiping Yu, Shunfeng Zhou, Shuting Pan, T.~Wang, Tao Yun, Tian Pei, Tianyu Sun, W.~L. Xiao, and Wangding Zeng.
\newblock Deepseek-v3 technical report.
\newblock \emph{CoRR}, abs/2412.19437, 2024.
\newblock \doi{10.48550/ARXIV.2412.19437}.
\newblock URL \url{https://doi.org/10.48550/arXiv.2412.19437}.

\bibitem[DeepSeek{-}AI et~al.(2025)DeepSeek{-}AI, Guo, Yang, Zhang, Song, Zhang, Xu, Zhu, Ma, Wang, Bi, Zhang, Yu, Wu, Wu, Gou, Shao, Li, Gao, Liu, Xue, Wang, Wu, Feng, Lu, Zhao, Deng, Zhang, Ruan, Dai, Chen, Ji, Li, Lin, Dai, Luo, Hao, Chen, Li, Zhang, Bao, Xu, Wang, Ding, Xin, Gao, Qu, Li, Guo, Li, Wang, Chen, Yuan, Qiu, Li, Cai, Ni, Liang, Chen, Dong, Hu, Gao, Guan, Huang, Yu, Wang, Zhang, Zhao, Wang, Zhang, Xu, Xia, Zhang, Zhang, Tang, Li, Wang, Li, Tian, Huang, Zhang, Wang, Chen, Du, Ge, Zhang, Pan, Wang, Chen, Jin, Chen, Lu, Zhou, Chen, Ye, Wang, Yu, Zhou, Pan, and Li]{guo2024deepseekr1}
DeepSeek{-}AI, Daya Guo, Dejian Yang, Haowei Zhang, Junxiao Song, Ruoyu Zhang, Runxin Xu, Qihao Zhu, Shirong Ma, Peiyi Wang, Xiao Bi, Xiaokang Zhang, Xingkai Yu, Yu~Wu, Z.~F. Wu, Zhibin Gou, Zhihong Shao, Zhuoshu Li, Ziyi Gao, Aixin Liu, Bing Xue, Bingxuan Wang, Bochao Wu, Bei Feng, Chengda Lu, Chenggang Zhao, Chengqi Deng, Chenyu Zhang, Chong Ruan, Damai Dai, Deli Chen, Dongjie Ji, Erhang Li, Fangyun Lin, Fucong Dai, Fuli Luo, Guangbo Hao, Guanting Chen, Guowei Li, H.~Zhang, Han Bao, Hanwei Xu, Haocheng Wang, Honghui Ding, Huajian Xin, Huazuo Gao, Hui Qu, Hui Li, Jianzhong Guo, Jiashi Li, Jiawei Wang, Jingchang Chen, Jingyang Yuan, Junjie Qiu, Junlong Li, J.~L. Cai, Jiaqi Ni, Jian Liang, Jin Chen, Kai Dong, Kai Hu, Kaige Gao, Kang Guan, Kexin Huang, Kuai Yu, Lean Wang, Lecong Zhang, Liang Zhao, Litong Wang, Liyue Zhang, Lei Xu, Leyi Xia, Mingchuan Zhang, Minghua Zhang, Minghui Tang, Meng Li, Miaojun Wang, Mingming Li, Ning Tian, Panpan Huang, Peng Zhang, Qiancheng Wang, Qinyu Chen, Qiushi Du, Ruiqi Ge,
  Ruisong Zhang, Ruizhe Pan, Runji Wang, R.~J. Chen, R.~L. Jin, Ruyi Chen, Shanghao Lu, Shangyan Zhou, Shanhuang Chen, Shengfeng Ye, Shiyu Wang, Shuiping Yu, Shunfeng Zhou, Shuting Pan, and S.~S. Li.
\newblock Deepseek-r1: Incentivizing reasoning capability in llms via reinforcement learning.
\newblock \emph{CoRR}, abs/2501.12948, 2025.
\newblock \doi{10.48550/ARXIV.2501.12948}.
\newblock URL \url{https://doi.org/10.48550/arXiv.2501.12948}.

\bibitem[Duan et~al.(2022)Duan, Yu, Tan, Zhu, and Tan]{duan2022embodiedsurvey}
Jiafei Duan, Samson Yu, Hui~Li Tan, Hongyuan Zhu, and Cheston Tan.
\newblock A survey of embodied {AI:} from simulators to research tasks.
\newblock \emph{{IEEE} Trans. Emerg. Top. Comput. Intell.}, 6\penalty0 (2):\penalty0 230--244, 2022.
\newblock \doi{10.1109/TETCI.2022.3141105}.
\newblock URL \url{https://doi.org/10.1109/TETCI.2022.3141105}.

\bibitem[Gibson(1979)]{Gibson1979GIBTEA}
James~J. Gibson.
\newblock \emph{The Ecological Approach to Visual Perception: Classic Edition}.
\newblock Houghton Mifflin, 1979.

\bibitem[Guo et~al.(2024)Guo, Chen, Wang, Chang, Pei, Chawla, Wiest, and Zhang]{guo2024agentsurvey}
Taicheng Guo, Xiuying Chen, Yaqi Wang, Ruidi Chang, Shichao Pei, Nitesh~V. Chawla, Olaf Wiest, and Xiangliang Zhang.
\newblock Large language model based multi-agents: {A} survey of progress and challenges.
\newblock In \emph{Proceedings of the Thirty-Third International Joint Conference on Artificial Intelligence, {IJCAI} 2024, Jeju, South Korea, August 3-9, 2024}, pages 8048--8057. ijcai.org, 2024.
\newblock URL \url{https://www.ijcai.org/proceedings/2024/890}.

\bibitem[Hu et~al.(2025)Hu, Zhang, Han, Jiang, Zhang, and Shum]{hu2025open}
Jingcheng Hu, Yinmin Zhang, Qi~Han, Daxin Jiang, Xiangyu Zhang, and Heung{-}Yeung Shum.
\newblock Open-reasoner-zero: An open source approach to scaling up reinforcement learning on the base model.
\newblock \emph{CoRR}, abs/2503.24290, 2025.
\newblock \doi{10.48550/ARXIV.2503.24290}.
\newblock URL \url{https://doi.org/10.48550/arXiv.2503.24290}.

\bibitem[Jia et~al.(2025)Jia, Li, Wang, Li, Hou, Dong, and Yu]{Jia2025ControllingLL}
Chengxing Jia, Ziniu Li, Pengyuan Wang, Yi-Chen Li, Zhenyu Hou, Yuxiao Dong, and Yang Yu.
\newblock Controlling large language model with latent actions.
\newblock \emph{ArXiv}, abs/2503.21383, 2025.
\newblock URL \url{https://api.semanticscholar.org/CorpusID:277349466}.

\bibitem[Jin et~al.(2025)Jin, Zeng, Yue, Wang, Zamani, and Han]{jin2025search}
Bowen Jin, Hansi Zeng, Zhenrui Yue, Dong Wang, Hamed Zamani, and Jiawei Han.
\newblock Search-r1: Training llms to reason and leverage search engines with reinforcement learning.
\newblock \emph{CoRR}, abs/2503.09516, 2025.
\newblock \doi{10.48550/ARXIV.2503.09516}.
\newblock URL \url{https://doi.org/10.48550/arXiv.2503.09516}.

\bibitem[Li et~al.(2025{\natexlab{a}})Li, Dong, Jin, Zhang, Zhou, Zhu, Zhang, and Dou]{li2025searcho1}
Xiaoxi Li, Guanting Dong, Jiajie Jin, Yuyao Zhang, Yujia Zhou, Yutao Zhu, Peitian Zhang, and Zhicheng Dou.
\newblock Search-o1: Agentic search-enhanced large reasoning models.
\newblock \emph{CoRR}, abs/2501.05366, 2025{\natexlab{a}}.
\newblock \doi{10.48550/ARXIV.2501.05366}.
\newblock URL \url{https://doi.org/10.48550/arXiv.2501.05366}.

\bibitem[Li et~al.(2025{\natexlab{b}})Li, Zhang, Zhang, Zhang, Liu, Yao, Xu, Zheng, Wang, Chen, Zhang, Yin, Dong, Guo, Song, and Liu]{li2025reasoningsurvey}
Zhong{-}Zhi Li, Duzhen Zhang, Ming{-}Liang Zhang, Jiaxin Zhang, Zengyan Liu, Yuxuan Yao, Haotian Xu, Junhao Zheng, Pei{-}Jie Wang, Xiuyi Chen, Yingying Zhang, Fei Yin, Jiahua Dong, Zhijiang Guo, Le~Song, and Cheng{-}Lin Liu.
\newblock From system 1 to system 2: {A} survey of reasoning large language models.
\newblock \emph{CoRR}, abs/2502.17419, 2025{\natexlab{b}}.
\newblock \doi{10.48550/ARXIV.2502.17419}.
\newblock URL \url{https://doi.org/10.48550/arXiv.2502.17419}.

\bibitem[Lightman et~al.(2024)Lightman, Kosaraju, Burda, Edwards, Baker, Lee, Leike, Schulman, Sutskever, and Cobbe]{lightman2024lets}
Hunter Lightman, Vineet Kosaraju, Yuri Burda, Harrison Edwards, Bowen Baker, Teddy Lee, Jan Leike, John Schulman, Ilya Sutskever, and Karl Cobbe.
\newblock Let's verify step by step.
\newblock In \emph{The Twelfth International Conference on Learning Representations, {ICLR} 2024, Vienna, Austria, May 7-11, 2024}. OpenReview.net, 2024.
\newblock URL \url{https://openreview.net/forum?id=v8L0pN6EOi}.

\bibitem[Lin et~al.(2025)Lin, Tang, Yao, Yin, Hu, Sun, and Chang]{Lin2025QLASSBL}
Zongyu Lin, Yao Tang, Xingcheng Yao, Da~Yin, Ziniu Hu, Yizhou Sun, and Kai-Wei Chang.
\newblock Qlass: Boosting language agent inference via q-guided stepwise search.
\newblock \emph{ArXiv}, abs/2502.02584, 2025.
\newblock URL \url{https://api.semanticscholar.org/CorpusID:276106886}.

\bibitem[Liu et~al.(2024)Liu, Chen, Bai, Li, Gao, and Lin]{liu2024embodiedsurvey}
Yang Liu, Weixing Chen, Yongjie Bai, Guanbin Li, Wen Gao, and Liang Lin.
\newblock Aligning cyber space with physical world: {A} comprehensive survey on embodied {AI}.
\newblock \emph{CoRR}, abs/2407.06886, 2024.
\newblock \doi{10.48550/ARXIV.2407.06886}.
\newblock URL \url{https://doi.org/10.48550/arXiv.2407.06886}.

\bibitem[Liu et~al.(2025)Liu, Peng, Zhong, Yue, Lu, Yu, and Jia]{liu2025segzero}
Yuqi Liu, Bohao Peng, Zhisheng Zhong, Zihao Yue, Fanbin Lu, Bei Yu, and Jiaya Jia.
\newblock Seg-zero: Reasoning-chain guided segmentation via cognitive reinforcement.
\newblock \emph{CoRR}, abs/2503.06520, 2025.
\newblock \doi{10.48550/ARXIV.2503.06520}.
\newblock URL \url{https://doi.org/10.48550/arXiv.2503.06520}.

\bibitem[Luo et~al.(2023)Luo, Sun, Xu, Zhao, Lou, Tao, Geng, Lin, Chen, and Zhang]{luo2023wizardmath}
Haipeng Luo, Qingfeng Sun, Can Xu, Pu~Zhao, Jianguang Lou, Chongyang Tao, Xiubo Geng, Qingwei Lin, Shifeng Chen, and Dongmei Zhang.
\newblock Wizardmath: Empowering mathematical reasoning for large language models via reinforced evol-instruct.
\newblock \emph{CoRR}, abs/2308.09583, 2023.
\newblock \doi{10.48550/ARXIV.2308.09583}.
\newblock URL \url{https://doi.org/10.48550/arXiv.2308.09583}.

\bibitem[Lyu et~al.(2025)Lyu, Gao, Gu, Zhang, Gao, Liu, Wang, Li, Zhao, Huang, Cao, Liu, Liu, Liu, Zhang, Lin, and Chen]{lyu2025oreal}
Chengqi Lyu, Songyang Gao, Yuzhe Gu, Wenwei Zhang, Jianfei Gao, Kuikun Liu, Ziyi Wang, Shuaibin Li, Qian Zhao, Haian Huang, Weihan Cao, Jiangning Liu, Hongwei Liu, Junnan Liu, Songyang Zhang, Dahua Lin, and Kai Chen.
\newblock Exploring the limit of outcome reward for learning mathematical reasoning.
\newblock \emph{CoRR}, abs/2502.06781, 2025.
\newblock \doi{10.48550/ARXIV.2502.06781}.
\newblock URL \url{https://doi.org/10.48550/arXiv.2502.06781}.

\bibitem[Min et~al.(2022)Min, Chaplot, Ravikumar, Bisk, and Salakhutdinov]{min2022film}
So~Yeon Min, Devendra~Singh Chaplot, Pradeep~Kumar Ravikumar, Yonatan Bisk, and Ruslan Salakhutdinov.
\newblock {FILM:} following instructions in language with modular methods.
\newblock In \emph{The Tenth International Conference on Learning Representations, {ICLR} 2022, Virtual Event, April 25-29, 2022}. OpenReview.net, 2022.
\newblock URL \url{https://openreview.net/forum?id=qI4542Y2s1D}.

\bibitem[Nottingham et~al.(2024)Nottingham, Majumder, Mishra, Singh, Clark, and Fox]{Nottingham2024sso}
Kolby Nottingham, Bodhisattwa~Prasad Majumder, Bhavana~Dalvi Mishra, Sameer Singh, Peter Clark, and Roy Fox.
\newblock Skill set optimization: Reinforcing language model behavior via transferable skills.
\newblock In \emph{Forty-first International Conference on Machine Learning, {ICML} 2024, Vienna, Austria, July 21-27, 2024}. OpenReview.net, 2024.
\newblock URL \url{https://openreview.net/forum?id=9laB7ytoMp}.

\bibitem[OpenAI(2023)]{openai2023gpt4}
OpenAI.
\newblock {GPT-4} technical report.
\newblock \emph{CoRR}, abs/2303.08774, 2023.
\newblock \doi{10.48550/ARXIV.2303.08774}.
\newblock URL \url{https://doi.org/10.48550/arXiv.2303.08774}.

\bibitem[Pan et~al.(2025)Pan, Liu, Wu, Liu, Zhu, Li, Chen, Ouyang, and Rueckert]{pan2025medvlmr1}
Jiazhen Pan, Che Liu, Junde Wu, Fenglin Liu, Jiayuan Zhu, Hongwei~Bran Li, Chen Chen, Cheng Ouyang, and Daniel Rueckert.
\newblock Medvlm-r1: Incentivizing medical reasoning capability of vision-language models (vlms) via reinforcement learning.
\newblock \emph{CoRR}, abs/2502.19634, 2025.
\newblock \doi{10.48550/ARXIV.2502.19634}.
\newblock URL \url{https://doi.org/10.48550/arXiv.2502.19634}.

\bibitem[Putta et~al.(2024)Putta, Mills, Garg, Motwani, Finn, Garg, and Rafailov]{putta2024agentq}
Pranav Putta, Edmund Mills, Naman Garg, Sumeet Motwani, Chelsea Finn, Divyansh Garg, and Rafael Rafailov.
\newblock Agent {Q:} advanced reasoning and learning for autonomous {AI} agents.
\newblock \emph{CoRR}, abs/2408.07199, 2024.
\newblock \doi{10.48550/ARXIV.2408.07199}.
\newblock URL \url{https://doi.org/10.48550/arXiv.2408.07199}.

\bibitem[Qiao et~al.(2024)Qiao, Fang, Zhang, Zhu, Chen, Deng, Jiang, Xie, Huang, and Chen]{qiao2024wkm}
Shuofei Qiao, Runnan Fang, Ningyu Zhang, Yuqi Zhu, Xiang Chen, Shumin Deng, Yong Jiang, Pengjun Xie, Fei Huang, and Huajun Chen.
\newblock Agent planning with world knowledge model.
\newblock In Amir Globersons, Lester Mackey, Danielle Belgrave, Angela Fan, Ulrich Paquet, Jakub~M. Tomczak, and Cheng Zhang, editors, \emph{Advances in Neural Information Processing Systems 38: Annual Conference on Neural Information Processing Systems 2024, NeurIPS 2024, Vancouver, BC, Canada, December 10 - 15, 2024}, 2024.
\newblock URL \url{http://papers.nips.cc/paper\_files/paper/2024/hash/d032263772946dd5026e7f3cd22bce5b-Abstract-Conference.html}.

\bibitem[Sarch et~al.(2024)Sarch, Jang, Tarr, Cohen, Marino, and Fragkiadaki]{Sarch2024ICAL}
Gabriel Sarch, Lawrence Jang, Michael~J. Tarr, William~W. Cohen, Kenneth Marino, and Katerina Fragkiadaki.
\newblock {VLM} agents generate their own memories: Distilling experience into embodied programs of thought.
\newblock In Amir Globersons, Lester Mackey, Danielle Belgrave, Angela Fan, Ulrich Paquet, Jakub~M. Tomczak, and Cheng Zhang, editors, \emph{Advances in Neural Information Processing Systems 38: Annual Conference on Neural Information Processing Systems 2024, NeurIPS 2024, Vancouver, BC, Canada, December 10 - 15, 2024}, 2024.
\newblock URL \url{http://papers.nips.cc/paper\_files/paper/2024/hash/8ac50fd0a4eeeb1f077b17bb7c5353c3-Abstract-Conference.html}.

\bibitem[Shao et~al.(2024)Shao, Wang, Zhu, Xu, Song, Zhang, Li, Wu, and Guo]{shao2024deepseekmath}
Zhihong Shao, Peiyi Wang, Qihao Zhu, Runxin Xu, Junxiao Song, Mingchuan Zhang, Y.~K. Li, Y.~Wu, and Daya Guo.
\newblock Deepseekmath: Pushing the limits of mathematical reasoning in open language models.
\newblock \emph{CoRR}, abs/2402.03300, 2024.
\newblock \doi{10.48550/ARXIV.2402.03300}.
\newblock URL \url{https://doi.org/10.48550/arXiv.2402.03300}.

\bibitem[Shinn et~al.(2023)Shinn, Cassano, Gopinath, Narasimhan, and Yao]{shinn2023reflexion}
Noah Shinn, Federico Cassano, Ashwin Gopinath, Karthik Narasimhan, and Shunyu Yao.
\newblock Reflexion: language agents with verbal reinforcement learning.
\newblock In Alice Oh, Tristan Naumann, Amir Globerson, Kate Saenko, Moritz Hardt, and Sergey Levine, editors, \emph{Advances in Neural Information Processing Systems 36: Annual Conference on Neural Information Processing Systems 2023, NeurIPS 2023, New Orleans, LA, USA, December 10 - 16, 2023}, 2023.
\newblock URL \url{http://papers.nips.cc/paper\_files/paper/2023/hash/1b44b878bb782e6954cd888628510e90-Abstract-Conference.html}.

\bibitem[Shridhar et~al.(2021)Shridhar, Yuan, C{\^{o}}t{\'{e}}, Bisk, Trischler, and Hausknecht]{mohit2021alfworld}
Mohit Shridhar, Xingdi Yuan, Marc{-}Alexandre C{\^{o}}t{\'{e}}, Yonatan Bisk, Adam Trischler, and Matthew~J. Hausknecht.
\newblock Alfworld: Aligning text and embodied environments for interactive learning.
\newblock In \emph{9th International Conference on Learning Representations, {ICLR} 2021, Virtual Event, Austria, May 3-7, 2021}. OpenReview.net, 2021.
\newblock URL \url{https://openreview.net/forum?id=0IOX0YcCdTn}.

\bibitem[Song et~al.(2024{\natexlab{a}})Song, Xiong, Zhao, Zhu, Wu, Wang, Li, Peng, and Li]{song2024agentbank}
Yifan Song, Weimin Xiong, Xiutian Zhao, Dawei Zhu, Wenhao Wu, Ke~Wang, Cheng Li, Wei Peng, and Sujian Li.
\newblock Agentbank: Towards generalized {LLM} agents via fine-tuning on 50000+ interaction trajectories.
\newblock In Yaser Al{-}Onaizan, Mohit Bansal, and Yun{-}Nung Chen, editors, \emph{Findings of the Association for Computational Linguistics: {EMNLP} 2024, Miami, Florida, USA, November 12-16, 2024}, pages 2124--2141. Association for Computational Linguistics, 2024{\natexlab{a}}.
\newblock URL \url{https://aclanthology.org/2024.findings-emnlp.116}.

\bibitem[Song et~al.(2024{\natexlab{b}})Song, Yin, Yue, Huang, Li, and Lin]{song2024eto}
Yifan Song, Da~Yin, Xiang Yue, Jie Huang, Sujian Li, and Bill~Yuchen Lin.
\newblock Trial and error: Exploration-based trajectory optimization for {LLM} agents.
\newblock \emph{CoRR}, abs/2403.02502, 2024{\natexlab{b}}.
\newblock \doi{10.48550/ARXIV.2403.02502}.
\newblock URL \url{https://doi.org/10.48550/arXiv.2403.02502}.

\bibitem[Sun et~al.(2023{\natexlab{a}})Sun, Zhuang, Kong, Dai, and Zhang]{sun2023adaplanner}
Haotian Sun, Yuchen Zhuang, Lingkai Kong, Bo~Dai, and Chao Zhang.
\newblock Adaplanner: Adaptive planning from feedback with language models.
\newblock In Alice Oh, Tristan Naumann, Amir Globerson, Kate Saenko, Moritz Hardt, and Sergey Levine, editors, \emph{Advances in Neural Information Processing Systems 36: Annual Conference on Neural Information Processing Systems 2023, NeurIPS 2023, New Orleans, LA, USA, December 10 - 16, 2023}, 2023{\natexlab{a}}.
\newblock URL \url{http://papers.nips.cc/paper\_files/paper/2023/hash/b5c8c1c117618267944b2617add0a766-Abstract-Conference.html}.

\bibitem[Sun et~al.(2023{\natexlab{b}})Sun, Zheng, Xie, Liu, Chu, Qiu, Xu, Ding, Li, Geng, Wu, Wang, Chen, Yin, Ren, Fu, He, Yuan, Liu, Liu, Li, Dong, Cheng, Zhang, Heng, Dai, Luo, Wang, Wen, Qiu, Guo, Xiong, Liu, and Li]{sun2023reasoning}
Jiankai Sun, Chuanyang Zheng, Enze Xie, Zhengying Liu, Ruihang Chu, Jianing Qiu, Jiaqi Xu, Mingyu Ding, Hongyang Li, Mengzhe Geng, Yue Wu, Wenhai Wang, Junsong Chen, Zhangyue Yin, Xiaozhe Ren, Jie Fu, Junxian He, Wu~Yuan, Qi~Liu, Xihui Liu, Yu~Li, Hao Dong, Yu~Cheng, Ming Zhang, Pheng{-}Ann Heng, Jifeng Dai, Ping Luo, Jingdong Wang, Ji{-}Rong Wen, Xipeng Qiu, Yike Guo, Hui Xiong, Qun Liu, and Zhenguo Li.
\newblock A survey of reasoning with foundation models.
\newblock \emph{CoRR}, abs/2312.11562, 2023{\natexlab{b}}.
\newblock \doi{10.48550/ARXIV.2312.11562}.
\newblock URL \url{https://doi.org/10.48550/arXiv.2312.11562}.

\bibitem[Team et~al.(2024)Team, Cai, Cao, Chen, Chen, Chen, Chen, Chen, Chen, Chen, Chu, Dong, Duan, Fan, Fei, Gao, Ge, Gu, Gu, Gui, Guo, Guo, He, Hu, Huang, Jiang, Jiao, Jin, Lei, Li, Li, Li, Li, Li, Li, Liu, Liu, Hong, Liu, Liu, Liu, Lv, Lv, Lv, Ma, Ma, Ma, Ning, Ouyang, Qiu, Qu, Shang, Shao, Song, Song, Sui, Sun, Sun, Tang, Wang, Wang, Wang, Wang, Wang, Wang, Wang, Wei, Weng, Wu, Xiong, Zhao, and et~al.]{cai24internlm2}
InternLM Team, Zheng Cai, Maosong Cao, Haojiong Chen, Kai Chen, Keyu Chen, Xin Chen, Xun Chen, Zehui Chen, Zhi Chen, Pei Chu, Xiaoyi Dong, Haodong Duan, Qi~Fan, Zhaoye Fei, Yang Gao, Jiaye Ge, Chenya Gu, Yuzhe Gu, Tao Gui, Aijia Guo, Qipeng Guo, Conghui He, Yingfan Hu, Ting Huang, Tao Jiang, Penglong Jiao, Zhenjiang Jin, Zhikai Lei, Jiaxing Li, Jingwen Li, Linyang Li, Shuaibin Li, Wei Li, Yining Li, Hongwei Liu, Jiangning Liu, Jiawei Hong, Kaiwen Liu, Kuikun Liu, Xiaoran Liu, Chengqi Lv, Haijun Lv, Kai Lv, Li~Ma, Runyuan Ma, Zerun Ma, Wenchang Ning, Linke Ouyang, Jiantao Qiu, Yuan Qu, Fukai Shang, Yunfan Shao, Demin Song, Zifan Song, Zhihao Sui, Peng Sun, Yu~Sun, Huanze Tang, Bin Wang, Guoteng Wang, Jiaqi Wang, Jiayu Wang, Rui Wang, Yudong Wang, Ziyi Wang, Xingjian Wei, Qizhen Weng, Fan Wu, Yingtong Xiong, Xiaomeng Zhao, and et~al.
\newblock Internlm2 technical report.
\newblock \emph{CoRR}, abs/2403.17297, 2024.
\newblock \doi{10.48550/ARXIV.2403.17297}.
\newblock URL \url{https://doi.org/10.48550/arXiv.2403.17297}.

\bibitem[Team et~al.(2025)Team, Du, Gao, Xing, Jiang, Chen, Li, Xiao, Du, Liao, Tang, Wang, Zhang, Yuan, Lu, Tang, Sung, Wei, Lai, Guo, Zhu, Ding, Hu, Yang, Zhang, Yao, Zhao, Lu, Li, Yu, Gao, Zheng, Yuan, Chen, Guo, Su, Wang, Zhao, Zhang, Liu, Yan, Wu, Shi, Ye, Yu, Dong, Zhang, Ma, Pan, Gong, Liu, Ma, Wei, Cao, Huang, Jiang, Gao, Xiong, He, Huang, Wu, He, Wei, Jia, Wu, Xu, Zu, Zhou, Pan, Charles, Li, Hu, Liu, Chen, Wang, Liu, Qin, Liu, Yang, Bao, Du, Wu, Wang, Zhou, Wang, Li, Zhu, Zhang, Wang, Yang, Huang, Huang, Xu, and Yang]{gao2025kimi}
Kimi Team, Angang Du, Bofei Gao, Bowei Xing, Changjiu Jiang, Cheng Chen, Cheng Li, Chenjun Xiao, Chenzhuang Du, Chonghua Liao, Chuning Tang, Congcong Wang, Dehao Zhang, Enming Yuan, Enzhe Lu, Fengxiang Tang, Flood Sung, Guangda Wei, Guokun Lai, Haiqing Guo, Han Zhu, Hao Ding, Hao Hu, Hao Yang, Hao Zhang, Haotian Yao, Haotian Zhao, Haoyu Lu, Haoze Li, Haozhen Yu, Hongcheng Gao, Huabin Zheng, Huan Yuan, Jia Chen, Jianhang Guo, Jianlin Su, Jianzhou Wang, Jie Zhao, Jin Zhang, Jingyuan Liu, Junjie Yan, Junyan Wu, Lidong Shi, Ling Ye, Longhui Yu, Mengnan Dong, Neo Zhang, Ningchen Ma, Qiwei Pan, Qucheng Gong, Shaowei Liu, Shengling Ma, Shupeng Wei, Sihan Cao, Siying Huang, Tao Jiang, Weihao Gao, Weimin Xiong, Weiran He, Weixiao Huang, Wenhao Wu, Wenyang He, Xianghui Wei, Xianqing Jia, Xingzhe Wu, Xinran Xu, Xinxing Zu, Xinyu Zhou, Xuehai Pan, Y.~Charles, Yang Li, Yangyang Hu, Yangyang Liu, Yanru Chen, Yejie Wang, Yibo Liu, Yidao Qin, Yifeng Liu, Ying Yang, Yiping Bao, Yulun Du, Yuxin Wu, Yuzhi Wang, Zaida Zhou,
  Zhaoji Wang, Zhaowei Li, Zhen Zhu, Zheng Zhang, Zhexu Wang, Zhilin Yang, Zhiqi Huang, Zihao Huang, Ziyao Xu, and Zonghan Yang.
\newblock Kimi k1.5: Scaling reinforcement learning with llms.
\newblock \emph{CoRR}, abs/2501.12599, 2025.
\newblock \doi{10.48550/ARXIV.2501.12599}.
\newblock URL \url{https://doi.org/10.48550/arXiv.2501.12599}.

\bibitem[Wang et~al.(2025{\natexlab{a}})Wang, Wang, Leong, and Li]{Wang2025STeCaST}
Hanlin Wang, Jian Wang, Chak~Tou Leong, and Wenjie Li.
\newblock Steca: Step-level trajectory calibration for llm agent learning.
\newblock \emph{ArXiv}, abs/2502.14276, 2025{\natexlab{a}}.
\newblock URL \url{https://api.semanticscholar.org/CorpusID:276482279}.

\bibitem[Wang et~al.(2024{\natexlab{a}})Wang, Hao, Dong, Zhang, Bao, Yang, and Wu]{wang2024oreo}
Huaijie Wang, Shibo Hao, Hanze Dong, Shenao Zhang, Yilin Bao, Ziran Yang, and Yi~Wu.
\newblock Offline reinforcement learning for {LLM} multi-step reasoning.
\newblock \emph{CoRR}, abs/2412.16145, 2024{\natexlab{a}}.
\newblock \doi{10.48550/ARXIV.2412.16145}.
\newblock URL \url{https://doi.org/10.48550/arXiv.2412.16145}.

\bibitem[Wang et~al.(2024{\natexlab{b}})Wang, Li, Han, Zhang, and Baldwin]{wang2024nat}
Renxi Wang, Haonan Li, Xudong Han, Yixuan Zhang, and Timothy Baldwin.
\newblock Learning from failure: Integrating negative examples when fine-tuning large language models as agents.
\newblock \emph{CoRR}, abs/2402.11651, 2024{\natexlab{b}}.
\newblock \doi{10.48550/ARXIV.2402.11651}.
\newblock URL \url{https://doi.org/10.48550/arXiv.2402.11651}.

\bibitem[Wang et~al.(2022)Wang, Jansen, C{\^{o}}t{\'{e}}, and Ammanabrolu]{wang2022scienceworld}
Ruoyao Wang, Peter~A. Jansen, Marc{-}Alexandre C{\^{o}}t{\'{e}}, and Prithviraj Ammanabrolu.
\newblock Scienceworld: Is your agent smarter than a 5th grader?
\newblock In Yoav Goldberg, Zornitsa Kozareva, and Yue Zhang, editors, \emph{Proceedings of the 2022 Conference on Empirical Methods in Natural Language Processing, {EMNLP} 2022, Abu Dhabi, United Arab Emirates, December 7-11, 2022}, pages 11279--11298. Association for Computational Linguistics, 2022.
\newblock \doi{10.18653/V1/2022.EMNLP-MAIN.775}.
\newblock URL \url{https://doi.org/10.18653/v1/2022.emnlp-main.775}.

\bibitem[Wang et~al.(2025{\natexlab{b}})Wang, Fei, Cheng, Zhang, Cai, Fu, and Qiu]{wang2025d2po}
Siyin Wang, Zhaoye Fei, Qinyuan Cheng, Shiduo Zhang, Panpan Cai, Jinlan Fu, and Xipeng Qiu.
\newblock World modeling makes a better planner: Dual preference optimization for embodied task planning.
\newblock \emph{CoRR}, abs/2503.10480, 2025{\natexlab{b}}.
\newblock \doi{10.48550/ARXIV.2503.10480}.
\newblock URL \url{https://doi.org/10.48550/arXiv.2503.10480}.

\bibitem[Wang et~al.(2023)Wang, Cai, Liu, Ma, and Liang]{wang2023dep}
Zihao Wang, Shaofei Cai, Anji Liu, Xiaojian Ma, and Yitao Liang.
\newblock Describe, explain, plan and select: Interactive planning with large language models enables open-world multi-task agents.
\newblock \emph{CoRR}, abs/2302.01560, 2023.
\newblock \doi{10.48550/ARXIV.2302.01560}.
\newblock URL \url{https://doi.org/10.48550/arXiv.2302.01560}.

\bibitem[Wei et~al.(2022)Wei, Wang, Schuurmans, Bosma, Ichter, Xia, Chi, Le, and Zhou]{wei2022cot}
Jason Wei, Xuezhi Wang, Dale Schuurmans, Maarten Bosma, Brian Ichter, Fei Xia, Ed~H. Chi, Quoc~V. Le, and Denny Zhou.
\newblock Chain-of-thought prompting elicits reasoning in large language models.
\newblock In Sanmi Koyejo, S.~Mohamed, A.~Agarwal, Danielle Belgrave, K.~Cho, and A.~Oh, editors, \emph{Advances in Neural Information Processing Systems 35: Annual Conference on Neural Information Processing Systems 2022, NeurIPS 2022, New Orleans, LA, USA, November 28 - December 9, 2022}, 2022.
\newblock URL \url{http://papers.nips.cc/paper\_files/paper/2022/hash/9d5609613524ecf4f15af0f7b31abca4-Abstract-Conference.html}.

\bibitem[Xi et~al.(2024)Xi, Ding, Chen, Hong, Guo, Wang, Yang, Liao, Guo, He, Gao, Chen, Zheng, Zou, Gui, Zhang, Qiu, Huang, Wu, and Jiang]{xi2024agentgym}
Zhiheng Xi, Yiwen Ding, Wenxiang Chen, Boyang Hong, Honglin Guo, Junzhe Wang, Dingwen Yang, Chenyang Liao, Xin Guo, Wei He, Songyang Gao, Lu~Chen, Rui Zheng, Yicheng Zou, Tao Gui, Qi~Zhang, Xipeng Qiu, Xuanjing Huang, Zuxuan Wu, and Yu{-}Gang Jiang.
\newblock Agentgym: Evolving large language model-based agents across diverse environments.
\newblock \emph{CoRR}, abs/2406.04151, 2024.
\newblock \doi{10.48550/ARXIV.2406.04151}.
\newblock URL \url{https://doi.org/10.48550/arXiv.2406.04151}.

\bibitem[Xi et~al.(2025)Xi, Chen, Guo, He, Ding, Hong, Zhang, Wang, Jin, Zhou, Zheng, Fan, Wang, Xiong, Zhou, Wang, Jiang, Zou, Liu, Yin, Dou, Weng, Qin, Zheng, Qiu, Huang, Zhang, and Gui]{xi2025agentsurvey}
Zhiheng Xi, Wenxiang Chen, Xin Guo, Wei He, Yiwen Ding, Boyang Hong, Ming Zhang, Junzhe Wang, Senjie Jin, Enyu Zhou, Rui Zheng, Xiaoran Fan, Xiao Wang, Limao Xiong, Yuhao Zhou, Weiran Wang, Changhao Jiang, Yicheng Zou, Xiangyang Liu, Zhangyue Yin, Shihan Dou, Rongxiang Weng, Wenjuan Qin, Yongyan Zheng, Xipeng Qiu, Xuanjing Huang, Qi~Zhang, and Tao Gui.
\newblock The rise and potential of large language model based agents: a survey.
\newblock \emph{Sci. China Inf. Sci.}, 68\penalty0 (2), 2025.
\newblock \doi{10.1007/S11432-024-4222-0}.
\newblock URL \url{https://doi.org/10.1007/s11432-024-4222-0}.

\bibitem[Xie et~al.(2025)Xie, Gao, Ren, Luo, Hong, Dai, Zhou, Qiu, Wu, and Luo]{xie2025logic}
Tian Xie, Zitian Gao, Qingnan Ren, Haoming Luo, Yuqian Hong, Bryan Dai, Joey Zhou, Kai Qiu, Zhirong Wu, and Chong Luo.
\newblock Logic-rl: Unleashing {LLM} reasoning with rule-based reinforcement learning.
\newblock \emph{CoRR}, abs/2502.14768, 2025.
\newblock \doi{10.48550/ARXIV.2502.14768}.
\newblock URL \url{https://doi.org/10.48550/arXiv.2502.14768}.

\bibitem[Xiong et~al.(2024{\natexlab{a}})Xiong, Song, Zhao, Wu, Wang, Wang, Li, Peng, and Li]{xiong-etal-2024-ipr}
Weimin Xiong, Yifan Song, Xiutian Zhao, Wenhao Wu, Xun Wang, Ke~Wang, Cheng Li, Wei Peng, and Sujian Li.
\newblock Watch every step! {LLM} agent learning via iterative step-level process refinement.
\newblock In Yaser Al-Onaizan, Mohit Bansal, and Yun-Nung Chen, editors, \emph{Proceedings of the 2024 Conference on Empirical Methods in Natural Language Processing}, pages 1556--1572, Miami, Florida, USA, November 2024{\natexlab{a}}. Association for Computational Linguistics.
\newblock \doi{10.18653/v1/2024.emnlp-main.93}.
\newblock URL \url{https://aclanthology.org/2024.emnlp-main.93/}.

\bibitem[Xiong et~al.(2024{\natexlab{b}})Xiong, Song, Zhao, Wu, Wang, Wang, Li, Peng, and Li]{xiong2024wes}
Weimin Xiong, Yifan Song, Xiutian Zhao, Wenhao Wu, Xun Wang, Ke~Wang, Cheng Li, Wei Peng, and Sujian Li.
\newblock Watch every step! {LLM} agent learning via iterative step-level process refinement.
\newblock In Yaser Al{-}Onaizan, Mohit Bansal, and Yun{-}Nung Chen, editors, \emph{Proceedings of the 2024 Conference on Empirical Methods in Natural Language Processing, {EMNLP} 2024, Miami, FL, USA, November 12-16, 2024}, pages 1556--1572. Association for Computational Linguistics, 2024{\natexlab{b}}.
\newblock URL \url{https://aclanthology.org/2024.emnlp-main.93}.

\bibitem[Yang et~al.(2024{\natexlab{a}})Yang, Yang, Zhang, Hui, Zheng, Yu, Li, Liu, Huang, Wei, Lin, Yang, Tu, Zhang, Yang, Yang, Zhou, Lin, Dang, Lu, Bao, Yang, Yu, Li, Xue, Zhang, Zhu, Men, Lin, Li, Xia, Ren, Ren, Fan, Su, Zhang, Wan, Liu, Cui, Zhang, and Qiu]{yang2024qwen2_5}
An~Yang, Baosong Yang, Beichen Zhang, Binyuan Hui, Bo~Zheng, Bowen Yu, Chengyuan Li, Dayiheng Liu, Fei Huang, Haoran Wei, Huan Lin, Jian Yang, Jianhong Tu, Jianwei Zhang, Jianxin Yang, Jiaxi Yang, Jingren Zhou, Junyang Lin, Kai Dang, Keming Lu, Keqin Bao, Kexin Yang, Le~Yu, Mei Li, Mingfeng Xue, Pei Zhang, Qin Zhu, Rui Men, Runji Lin, Tianhao Li, Tingyu Xia, Xingzhang Ren, Xuancheng Ren, Yang Fan, Yang Su, Yichang Zhang, Yu~Wan, Yuqiong Liu, Zeyu Cui, Zhenru Zhang, and Zihan Qiu.
\newblock Qwen2.5 technical report.
\newblock \emph{CoRR}, abs/2412.15115, 2024{\natexlab{a}}.
\newblock \doi{10.48550/ARXIV.2412.15115}.
\newblock URL \url{https://doi.org/10.48550/arXiv.2412.15115}.

\bibitem[Yang et~al.(2024{\natexlab{b}})Yang, Zhao, Gu, and Zhou]{yang2024cops}
Chen Yang, Chenyang Zhao, Quanquan Gu, and Dongruo Zhou.
\newblock Cops: Empowering {LLM} agents with provable cross-task experience sharing.
\newblock \emph{CoRR}, abs/2410.16670, 2024{\natexlab{b}}.
\newblock \doi{10.48550/ARXIV.2410.16670}.
\newblock URL \url{https://doi.org/10.48550/arXiv.2410.16670}.

\bibitem[Yao et~al.(2023)Yao, Zhao, Yu, Du, Shafran, Narasimhan, and Cao]{yao2023react}
Shunyu Yao, Jeffrey Zhao, Dian Yu, Nan Du, Izhak Shafran, Karthik~R. Narasimhan, and Yuan Cao.
\newblock React: Synergizing reasoning and acting in language models.
\newblock In \emph{The Eleventh International Conference on Learning Representations, {ICLR} 2023, Kigali, Rwanda, May 1-5, 2023}. OpenReview.net, 2023.
\newblock URL \url{https://openreview.net/forum?id=WE\_vluYUL-X}.

\bibitem[Yao et~al.(2024)Yao, Heinecke, Niebles, Liu, Feng, Xue, N., Chen, Zhang, Arpit, Xu, Mui, Wang, Xiong, and Savarese]{yao2024retroformer}
Weiran Yao, Shelby Heinecke, Juan~Carlos Niebles, Zhiwei Liu, Yihao Feng, Le~Xue, Rithesh~R. N., Zeyuan Chen, Jianguo Zhang, Devansh Arpit, Ran Xu, Phil Mui, Huan Wang, Caiming Xiong, and Silvio Savarese.
\newblock Retroformer: Retrospective large language agents with policy gradient optimization.
\newblock In \emph{The Twelfth International Conference on Learning Representations, {ICLR} 2024, Vienna, Austria, May 7-11, 2024}. OpenReview.net, 2024.
\newblock URL \url{https://openreview.net/forum?id=KOZu91CzbK}.

\bibitem[Yin et~al.(2024)Yin, Brahman, Ravichander, Chandu, Chang, Choi, and Lin]{yin2024lumos}
Da~Yin, Faeze Brahman, Abhilasha Ravichander, Khyathi~Raghavi Chandu, Kai{-}Wei Chang, Yejin Choi, and Bill~Yuchen Lin.
\newblock Agent lumos: Unified and modular training for open-source language agents.
\newblock In Lun{-}Wei Ku, Andre Martins, and Vivek Srikumar, editors, \emph{Proceedings of the 62nd Annual Meeting of the Association for Computational Linguistics (Volume 1: Long Papers), {ACL} 2024, Bangkok, Thailand, August 11-16, 2024}, pages 12380--12403. Association for Computational Linguistics, 2024.
\newblock \doi{10.18653/V1/2024.ACL-LONG.670}.
\newblock URL \url{https://doi.org/10.18653/v1/2024.acl-long.670}.

\bibitem[Ying et~al.(2024)Ying, Zhang, Li, Zhou, Shao, Fei, Ma, Hong, Liu, Wang, Wang, Wu, Li, Zhou, Liu, Zhang, Zhang, Yan, Qiu, Wang, Chen, and Lin]{ying2024internlm-math}
Huaiyuan Ying, Shuo Zhang, Linyang Li, Zhejian Zhou, Yunfan Shao, Zhaoye Fei, Yichuan Ma, Jiawei Hong, Kuikun Liu, Ziyi Wang, Yudong Wang, Zijian Wu, Shuaibin Li, Fengzhe Zhou, Hongwei Liu, Songyang Zhang, Wenwei Zhang, Hang Yan, Xipeng Qiu, Jiayu Wang, Kai Chen, and Dahua Lin.
\newblock Internlm-math: Open math large language models toward verifiable reasoning.
\newblock \emph{CoRR}, abs/2402.06332, 2024.
\newblock \doi{10.48550/ARXIV.2402.06332}.
\newblock URL \url{https://doi.org/10.48550/arXiv.2402.06332}.

\bibitem[Yuan et~al.(2025)Yuan, Chen, Xi, Ye, Du, and Chen]{yuan2025agentr}
Siyu Yuan, Zehui Chen, Zhiheng Xi, Junjie Ye, Zhengyin Du, and Jiecao Chen.
\newblock Agent-r: Training language model agents to reflect via iterative self-training.
\newblock \emph{CoRR}, abs/2501.11425, 2025.
\newblock \doi{10.48550/ARXIV.2501.11425}.
\newblock URL \url{https://doi.org/10.48550/arXiv.2501.11425}.

\bibitem[Zeng et~al.(2024{\natexlab{a}})Zeng, Liu, Lu, Wang, Liu, Dong, and Tang]{zeng-etal-2024-agenttuning}
Aohan Zeng, Mingdao Liu, Rui Lu, Bowen Wang, Xiao Liu, Yuxiao Dong, and Jie Tang.
\newblock {A}gent{T}uning: Enabling generalized agent abilities for {LLM}s.
\newblock In Lun-Wei Ku, Andre Martins, and Vivek Srikumar, editors, \emph{Findings of the Association for Computational Linguistics: ACL 2024}, pages 3053--3077, Bangkok, Thailand, August 2024{\natexlab{a}}. Association for Computational Linguistics.
\newblock \doi{10.18653/v1/2024.findings-acl.181}.
\newblock URL \url{https://aclanthology.org/2024.findings-acl.181/}.

\bibitem[Zeng et~al.(2024{\natexlab{b}})Zeng, Liu, Lu, Wang, Liu, Dong, and Tang]{zeng2024agenttuning}
Aohan Zeng, Mingdao Liu, Rui Lu, Bowen Wang, Xiao Liu, Yuxiao Dong, and Jie Tang.
\newblock Agenttuning: Enabling generalized agent abilities for llms.
\newblock In Lun{-}Wei Ku, Andre Martins, and Vivek Srikumar, editors, \emph{Findings of the Association for Computational Linguistics, {ACL} 2024, Bangkok, Thailand and virtual meeting, August 11-16, 2024}, pages 3053--3077. Association for Computational Linguistics, 2024{\natexlab{b}}.
\newblock \doi{10.18653/V1/2024.FINDINGS-ACL.181}.
\newblock URL \url{https://doi.org/10.18653/v1/2024.findings-acl.181}.

\bibitem[Zhai et~al.(2025)Zhai, Yang, Xu, Feng, Yang, Ding, and Wang]{zhai2025enhancing}
Yuanzhao Zhai, Tingkai Yang, Kele Xu, Dawei Feng, Cheng Yang, Bo~Ding, and Huaimin Wang.
\newblock Enhancing decision-making for {LLM} agents via step-level q-value models.
\newblock In Toby Walsh, Julie Shah, and Zico Kolter, editors, \emph{AAAI-25, Sponsored by the Association for the Advancement of Artificial Intelligence, February 25 - March 4, 2025, Philadelphia, PA, {USA}}, pages 27161--27169. {AAAI} Press, 2025.
\newblock \doi{10.1609/AAAI.V39I25.34924}.
\newblock URL \url{https://doi.org/10.1609/aaai.v39i25.34924}.

\bibitem[Zhao et~al.(2023)Zhao, Lee, and Hsu]{zhao2023ck}
Zirui Zhao, Wee~Sun Lee, and David Hsu.
\newblock Large language models as commonsense knowledge for large-scale task planning.
\newblock In Alice Oh, Tristan Naumann, Amir Globerson, Kate Saenko, Moritz Hardt, and Sergey Levine, editors, \emph{Advances in Neural Information Processing Systems 36: Annual Conference on Neural Information Processing Systems 2023, NeurIPS 2023, New Orleans, LA, USA, December 10 - 16, 2023}, 2023.
\newblock URL \url{http://papers.nips.cc/paper\_files/paper/2023/hash/65a39213d7d0e1eb5d192aa77e77eeb7-Abstract-Conference.html}.

\bibitem[Zheng et~al.(2025)Zheng, Fu, Hu, Cai, Ye, Lu, and Liu]{zheng2025deepresearcher}
Yuxiang Zheng, Dayuan Fu, Xiangkun Hu, Xiaojie Cai, Lyumanshan Ye, Pengrui Lu, and Pengfei Liu.
\newblock Deepresearcher: Scaling deep research via reinforcement learning in real-world environments, 2025.
\newblock URL \url{https://arxiv.org/abs/2504.03160}.

\bibitem[Zhou et~al.(2023)Zhou, Jiang, Li, Wu, Wang, Qiu, Zhang, Chen, Wu, Wang, Zhu, Chen, Zhang, Zhang, Chen, Cui, and Sachan]{zhou2024agentsurvey}
Wangchunshu Zhou, Yuchen~Eleanor Jiang, Long Li, Jialong Wu, Tiannan Wang, Shi Qiu, Jintian Zhang, Jing Chen, Ruipu Wu, Shuai Wang, Shiding Zhu, Jiyu Chen, Wentao Zhang, Ningyu Zhang, Huajun Chen, Peng Cui, and Mrinmaya Sachan.
\newblock Agents: An open-source framework for autonomous language agents.
\newblock \emph{CoRR}, abs/2309.07870, 2023.
\newblock \doi{10.48550/ARXIV.2309.07870}.
\newblock URL \url{https://doi.org/10.48550/arXiv.2309.07870}.

\bibitem[Zhu et~al.(2025)Zhu, Qiao, Ou, Deng, Lyu, Shen, Liang, Gu, Chen, and Zhang]{zhu-etal-2025-knowagent}
Yuqi Zhu, Shuofei Qiao, Yixin Ou, Shumin Deng, Shiwei Lyu, Yue Shen, Lei Liang, Jinjie Gu, Huajun Chen, and Ningyu Zhang.
\newblock {K}now{A}gent: Knowledge-augmented planning for {LLM}-based agents.
\newblock In Luis Chiruzzo, Alan Ritter, and Lu~Wang, editors, \emph{Findings of the Association for Computational Linguistics: NAACL 2025}, pages 3709--3732, Albuquerque, New Mexico, April 2025. Association for Computational Linguistics.
\newblock ISBN 979-8-89176-195-7.
\newblock URL \url{https://aclanthology.org/2025.findings-naacl.205/}.

\end{thebibliography}

\newpage

\appendix

\section{Pseudo Code of Group rollout of interaction with environment}\label{sec:rollout_code}

\begin{algorithm}[h]
  \caption{Group rollout of interaction with environment}\label{alg:rollout}
  \textbf{Input}: Initial policy $\pi_{\theta_0}$, dataset $\mathcal{D} = \{(q, e): q \in \mathcal{Q}, e \in \mathcal{E}\}$, where $\mathcal{Q}$ is the set of instructions and $\mathcal{E}$ is the set of environments with the group size set to $n$.
  \begin{algorithmic}[1]
    \State Freeze old policy $\pi_{\theta_{\text{old}}} \gets \pi_\theta$
    \State Sample batch $\mathcal{B} = \{(q_i, e_i)\}_{i=0}^K \sim \mathcal{D}$
    \State Initialize global trajectories buffer $\mathcal{T} \gets \emptyset$
    \For{$(q_i, e_i) \in \mathcal{B}$}  \Comment{Process tasks}
      \State Create $n$ environment instance $\{e_j\}^n$ copy from $e_i$
      \State Initialize $n$ trajectories buffer $\{\tau_j\}^n \gets \emptyset$ for environment instance $\{e_j\}^n$
      \For{$j = 1$ to $n$}
        \State $o_{j,0} \gets e_{j}.\text{reset}()$ 
        \State $\tau_j \gets (q_i, o_{j,0})$
      \EndFor
      \For{$k=1$ to $max\_steps$}  \Comment{Environment interaction}
        \State Generate response $\{m_{j, k} \sim \pi_{\theta_{\text{old}}}(\cdot|\tau_{j})\}^n$ for each trajectories $\tau_j$
        \State Extract actions $\{a_{{j}, k}\}^{n}$ from responses $\{m_{j, k}\}^{n}$
        \State Get the observation $\{o_{j, k}, \text{done}_{j, k} \gets e_{i_j}.\text{step}(a_{{j}, k})\}^{n}$
        \State Update trajectories $\{\tau_{j}\ \gets \tau_{j} \oplus (m_{j,k}, o_{j,k})\}^n$
        \If{all $\{\text{done}_{j, k}\}^n$} 
          \State \textbf{break}
        \EndIf
      \EndFor
      \State Save trajectories $\mathcal{T} \gets \mathcal{T} \cup \{\tau_{j}\}^n$
    \EndFor
  \end{algorithmic}
  \textbf{Output}: The grouped trajectories $\mathcal{T}$
\end{algorithm}

% 1) Our experiments were conducted in pure text environments, but our approach is generalizable to multimodal settings. In the future, we plan to expand our methods and extend them to the vision domain. 2) The computational resources required for our approach may limit its accessibility for researchers with limited computing capacity. Addressing these efficiency challenges represents an important direction for future work.

\section{Evaluation and Benchmark Details}
\label{app: benchamrk}

\textbf{ALFWorld~\citep{mohit2021alfworld}} ALFWorld encompasses six categories of planning tasks set primarily in home environments, covering not only basic object manipulation (such as pick and place) but also tasks requiring complex interaction sequences. For example, the heating task requires models to first identify target objects, move them to heating devices~(like microwave), execute the heating operation, and finally place them in designated locations to complete the task. 

Following prior research~\citep{yao2023react,song2024eto}, we evaluate model performance under two conditions: seen and unseen scenarios. Seen scenarios consist of task instances from rooms encountered during training, unseen scenarios comprise task instances from entirely new rooms with container arrangements and scene organizations distinctly different from training tasks, designed to evaluate the model's zero-shot generalization capabilities.

\textbf{ScienceWorld~\citep{wang2022scienceworld}} ScienceWorld presents a more challenging benchmark, requiring models to complete scientific experiments in a highly interactive environment. This environment contains approximately ten interconnected rooms, over 200 object types, and 25 executable actions. Compared to ALFWorld, ScienceWorld integrates a more sophisticated physics simulation engine, including thermodynamic and electrical systems, which places more rigorous demands on a model's planning capabilities and causal reasoning.

Following prior research~\citep{yao2023react,song2024eto}, we assessed model performance under two conditions: seen and unseen scenarios. Seen scenarios consist of task instances from rooms encountered during training, including the same task types, objects, containers, and room layouts, but with variations in object positions, quantities, and visual features (e.g., two blue pencils on a shelf instead of three red pencils in a drawer as seen in training data). Unseen scenarios comprise task instances from entirely new rooms with container arrangements and scene organizations distinctly different from training tasks, designed to evaluate the model's zero-shot generalization capabilities. During evaluating ScienceWorld, we maintained the ``easy'' setup consistent with previous work~\citep{song2024eto}, enabling all five assistive features: instant teleportation, pre‑opened containers, self‑watering flower pots, and two additional aids that remove similar micro‑action bottlenecks, thus reducing the burden of low-level operations and focusing the evaluation on high-level planning abilities. We also adopted the same balancing strategy as previous work~\citep{song2024eto}, implementing fair sampling at the task level.

\begin{figure}[h]
\begin{AIbox}{ALFWorld Templates:}

% {\color{red}\bf System Prompt:}\\
\textcolor{blue}{\emph{Environment description}}\\
You are an intelligent agent in a household environment and your target is to perform actions to complete the task goal. At the beginning of your interactions, you will be given the detailed description of the current environment and your goal to accomplish.\\

For each of your turn, you will be given the observation of the last turn. You should first think about the current condition and plan for your future actions, and then output your action in this turn. Your output must strictly follow this format:
\textbf{Thought:} <your thoughts> ; \textbf{Action:} <your next action>.\\

\textcolor{blue}{\emph{Available actions \& required format}}\\
The available actions are:\\
1.\,\texttt{go to (receptacle)}\\
2.\,\texttt{open (receptacle)}\\
\quad$\vdots$\\
13.\,\texttt{inventory:check your current inventory}
   \hfill\textcolor{orange}{\emph{Explanations for abstract actions}}\\
14.\,\texttt{done:Indicate that you believe the task is complete}\\
Where \texttt{(object)} refers to manipulable objects and \texttt{(receptacle)} refers to receptacles or locations.\\

\textcolor{blue}{\emph{Avoid hallucination}}\\
After your each turn, the environment will give you immediate feedback based on which you plan your next few steps, if the environment output: Nothing happens, that means the previous action is invalid and you should try more options.\\
You can only hold one object at a time. Before taking a new object, make sure you have placed down any object you are currently holding.\\
You should not assume or anticipate the feedback. Even if you have planned multiple steps ahead, you should only execute one action at a time.\\
Do not proceed with any further exploration or actions until you receive the feedback from the environment after your action.\\

\textcolor{blue}{\emph{Output template}}\\
Your response should use the following format:\\
\textbf{Thought:} <your thoughts>\\
\textbf{Action:} <your next action>

\end{AIbox}
\caption{Prompt for ALFWorld}\label{alfworld_prompt}
\end{figure}

\section{Prompt Design}\label{app: prompt_design}

\begin{figure}[ht]
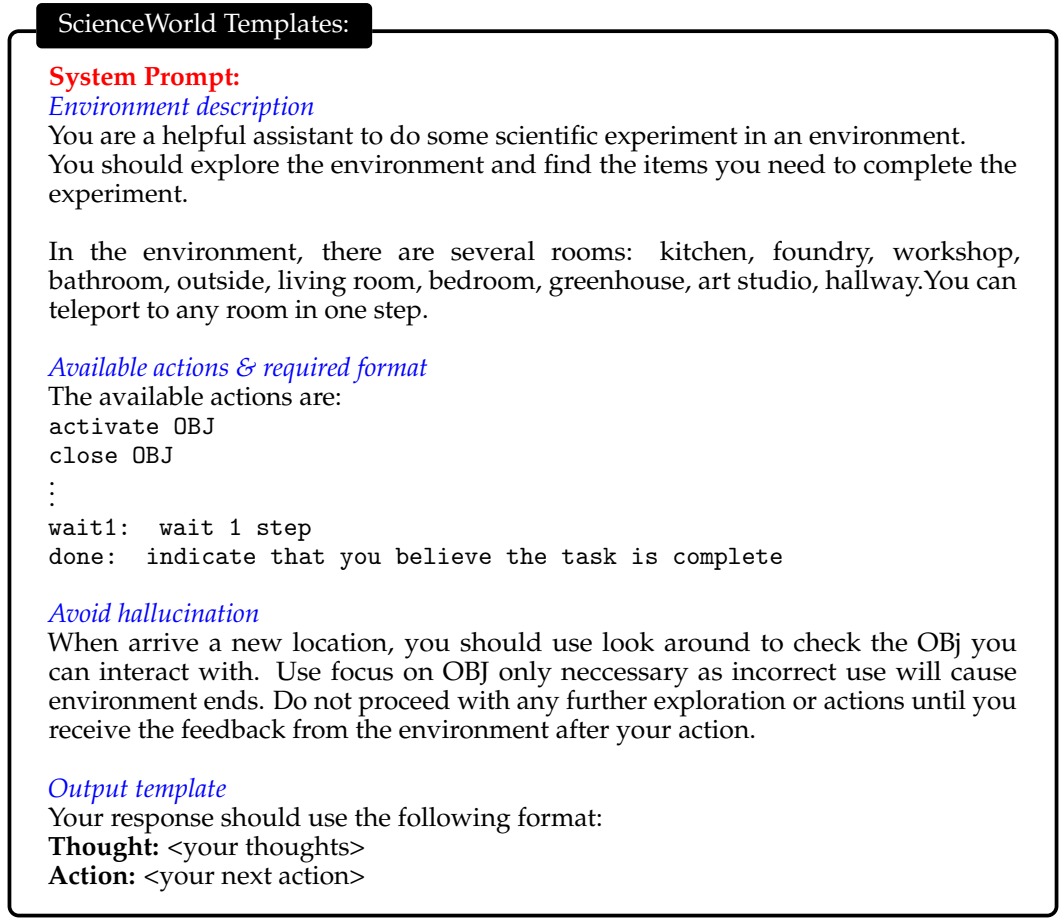

\begin{AIbox}{ScienceWorld Templates:}

{\color{red}\bf System Prompt:}\\
\textcolor{blue}{\emph{Environment description}}\\
You are a helpful assistant to do some scientific experiment in an environment.\\
You should explore the environment and find the items you need to complete the experiment.\\
\\
In the environment, there are several rooms: kitchen, foundry, workshop, bathroom, outside, living room, bedroom, greenhouse, art studio, hallway.You can teleport to any room in one step. \\

\textcolor{blue}{\emph{Available actions \& required format}}\\
The available actions are:\\
\texttt{activate OBJ} \\
\texttt{close OBJ} \\
\quad$\vdots$\\
\texttt{wait1: wait 1 step} \\
\texttt{done: indicate that you believe the task is complete} \\

\textcolor{blue}{\emph{Avoid hallucination}}\\
When arrive a new location, you should use look around to check the OBj you can interact with.
Use focus on OBJ only neccessary as incorrect use will cause environment ends.
Do not proceed with any further exploration or actions until you receive the feedback from the environment after your action.\\

\textcolor{blue}{\emph{Output template}}\\
Your response should use the following format:\\
\textbf{Thought:} <your thoughts>\\
\textbf{Action:} <your next action>

\end{AIbox}
\caption{Prompt for ScienceWorld}\label{fig:scienceworld_prompt}
\end{figure}

In our research, prompt engineering accomplishes three interconnected objectives: (i) enumerating permissible action, guiding the model to execute effective commands; (ii) mitigating common hallucinations, preventing the model from falling into blind trial-and-error loops; and (iii) enforcing a concise ReAct paradigm that clearly demonstrates reasoning chains and actions, facilitating straightforward extraction of selected operations. Following~\citep{qiao2024wkm}, we enhanced the system instructions to explicitly highlight the allowable action set, and required output format, while specifically addressing common hallucination patterns. This approach significantly increasing the probability of obtaining initial successful trajectories necessary for guided learning. The prompt is detailed in Figure~\ref{alfworld_prompt} and Figure~\ref{fig:scienceworld_prompt}.

\section{More experimental results}

\begin{figure}[t]
    \centering
    \includegraphics[width=\linewidth]{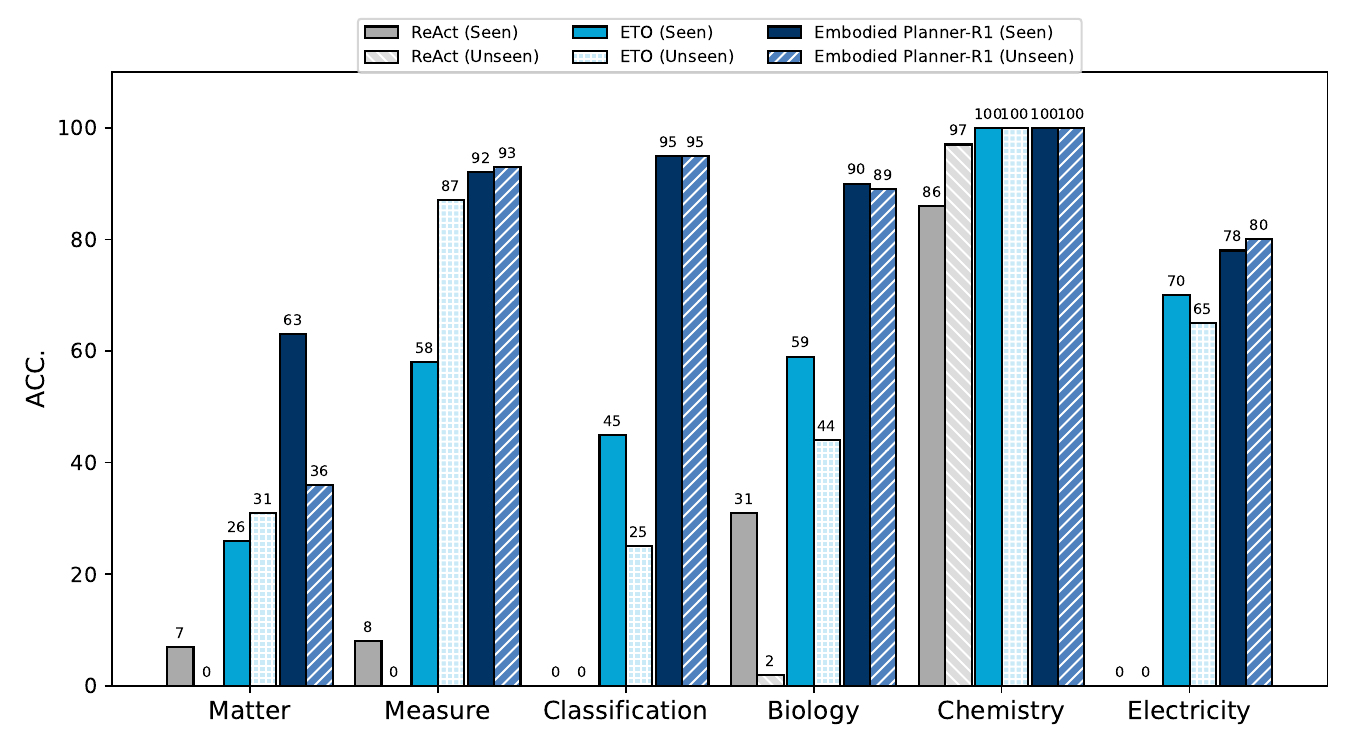}
    \label{fig:sciworld_seen}
    \caption{Completion rates of Qwen2.5-7B-Instruct on various ScienceWorld tasks, covering both seen and unseen scenarios. We compare the prompt-based method ReAct with the training-based method
ETO and \framework.}
    \label{fig:sciworld_table_combined}
\end{figure}

\begin{figure}[t]
    \begin{subfigure}[b]{0.49\linewidth}
        \centering
        \includegraphics[width=\linewidth]{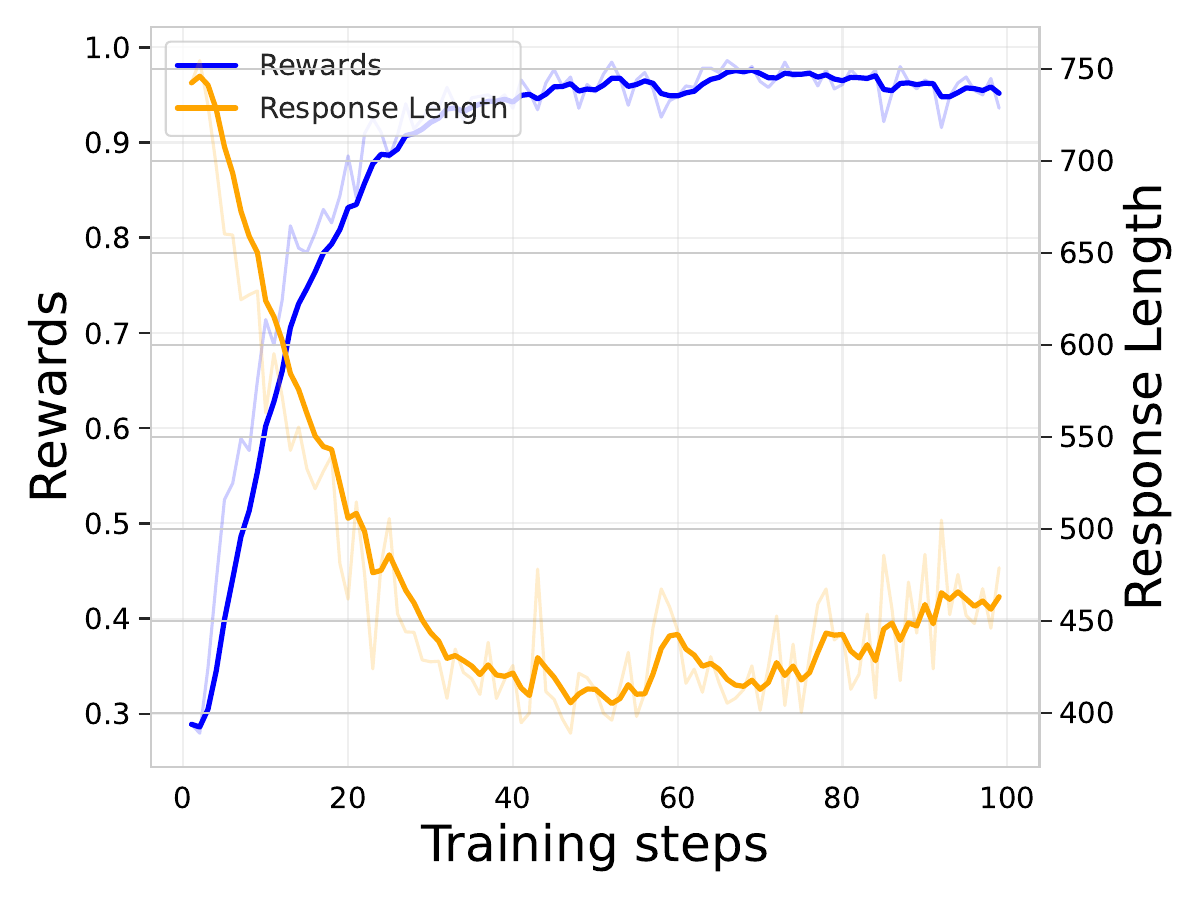}
        \caption{Rewards and response length}
        \label{fig:alfworl_progress}
    \end{subfigure}
    \begin{subfigure}[b]{0.49\linewidth}
        \centering
        \includegraphics[width=\linewidth]{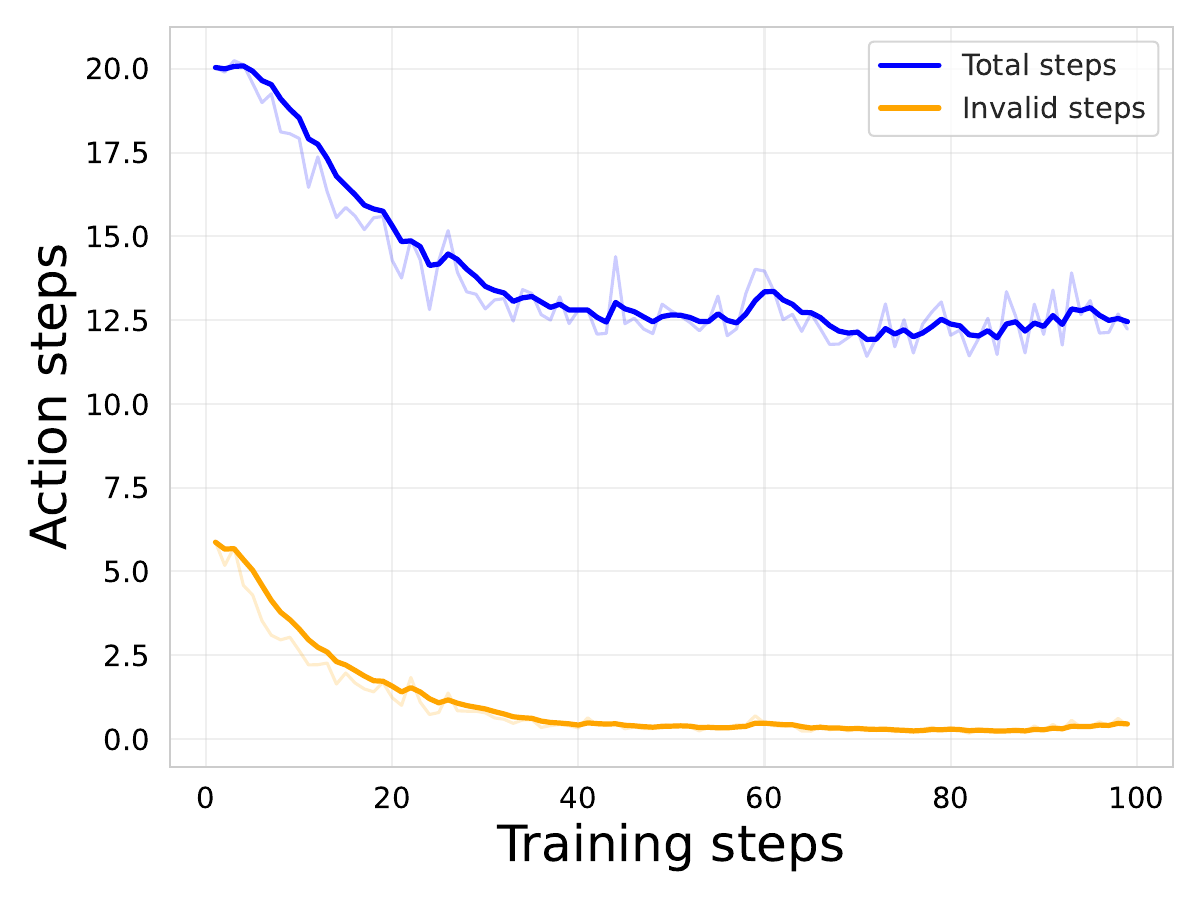}
        \caption{Total and invalid action steps}
        \label{fig:alfworld_steps}
    \end{subfigure}
    \caption{Training progress of \framework\ in ALFWorld: (a) shows the rewards (task completion rate) and total response length in multi-round exploration; (b) illustrates the decreasing trends in both total and invalid action steps during training.}\label{fig:alfworl_combined}
    \vspace{-1em}
    % \label{fig:combined}
\end{figure}

\begin{table}[!h]
    \centering
    \caption{ }\label{tab:appendix_results}%
    \begin{tabular}{lcc}
    \toprule
    Method & ALFWorld & ScienceWorld \\
    \midrule
    CoLA* \citep{Jia2025ControllingLL} &  76.3 & 25.1 \\
    IPR* \citep{xiong-etal-2024-ipr} &    72.5  & - \\
    STeCa*  \citep{Wang2025STeCaST} &   75.2 &  - \\
    Agent-R* \citep{yuan2025agentr} & - &  70.23 \\
    OREO* \citep{wang2024oreo} &  79.9 &  - \\
    DEP* \citep{wang2023dep} &  76.0 &  - \\
    CoPS* \citep{yang2024cops} & 94.0 &  - \\
    AdaPlanner* \citep{sun2023adaplanner} &  91.79 &  - \\
    AgentPRM* \citep{choud2025agentprm} &  91.0 &  - \\
    AgentLM* \citep{zeng-etal-2024-agenttuning} &  86.0 & 20.8 \\
    QLASS* \citep{Lin2025QLASSBL} &  80.4 &  70.9 \\
    KnowAgent* \citep{zhu-etal-2025-knowagent} &  64.70 & 53.93 \\
    WKM* \citep{qiao2024wkm}  &  67.25  & 53.93 \\
    \framework ~(Ours)  & \textbf{97.78} & \textbf{79.92} \\
    \bottomrule
    \end{tabular}%
\end{table}%

Figure~\ref{fig:sciworld_combined} illustrates the metrics of the \framework\ during the training process in ALFWorld, which yields results similar to those in ScienceWorld. The reward curve exhibits a similar upward trend. Meanwhile, the response length significantly decreases from the outset, dropping from approximately 750 tokens initially to around 450 tokens by the end. The relationship between response length and action steps is also evident in Figure~\ref{fig:sciworld_steps}, where total actions, invalid actions, and response length show a consistent downward trend. Due to the reduced difficulty compared to ScienceWorld, invalid steps have been almost entirely eliminated in the final results.

In Figure~\ref{fig:sciworld_table_combined}, we present the task completion performance of our trained agents in ScienceWorld, alongside a comparison with ReAct and ETO. This comparison underscores the effectiveness and generalization capabilities of our \framework. While prompt-based and previous training-based methods often exhibit strong performance on tasks that emphasize common sense reasoning (e.g., chemistry), their success rates plummet when faced with tasks that heavily rely on environmental interaction (e.g., classification). In fact, many of these tasks remain unsolvable for prompt-based approaches. In contrast, our method demonstrates exceptional completion rates across both seen and unseen tasks, highlighting its robustness and adaptability in complex and varied scenarios.

\newpage

\section{Case Study: Model's Knowledge about Environment}\label{app: case study}

\begin{figure}[h]
    \centering
    \resizebox{.8\textwidth}{!}{
    \includegraphics{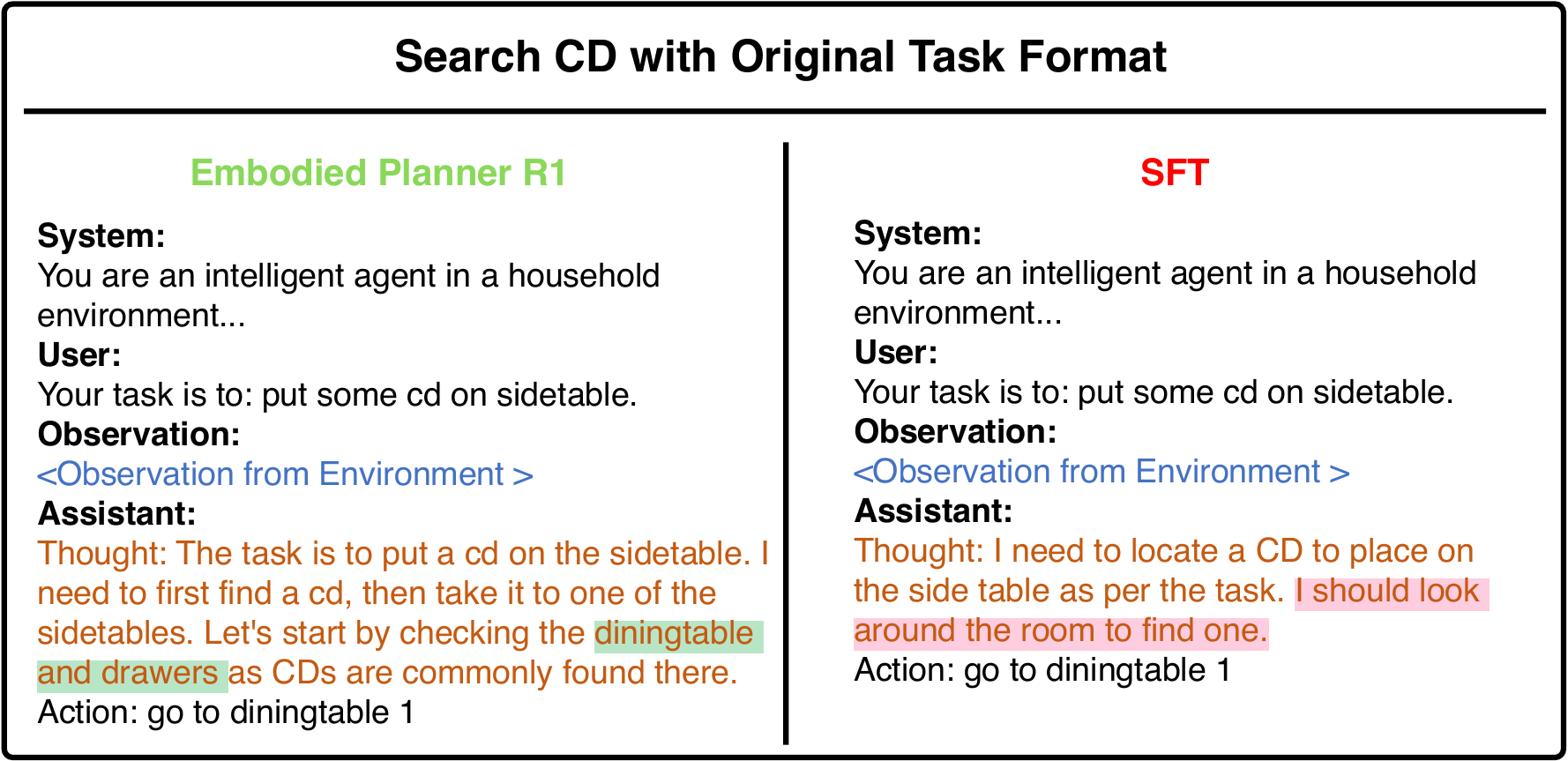}
    }
    \caption{Comparison between \framework\ and SFT using the original task format,where \framework\ pointed out the possible position of CD and SFT exhibited inconsistency between thought and action .}
    \label{fig:case-study-ori}
\end{figure}

\begin{figure}[th]
    \centering
    \resizebox{.8\textwidth}{!}{
    \includegraphics{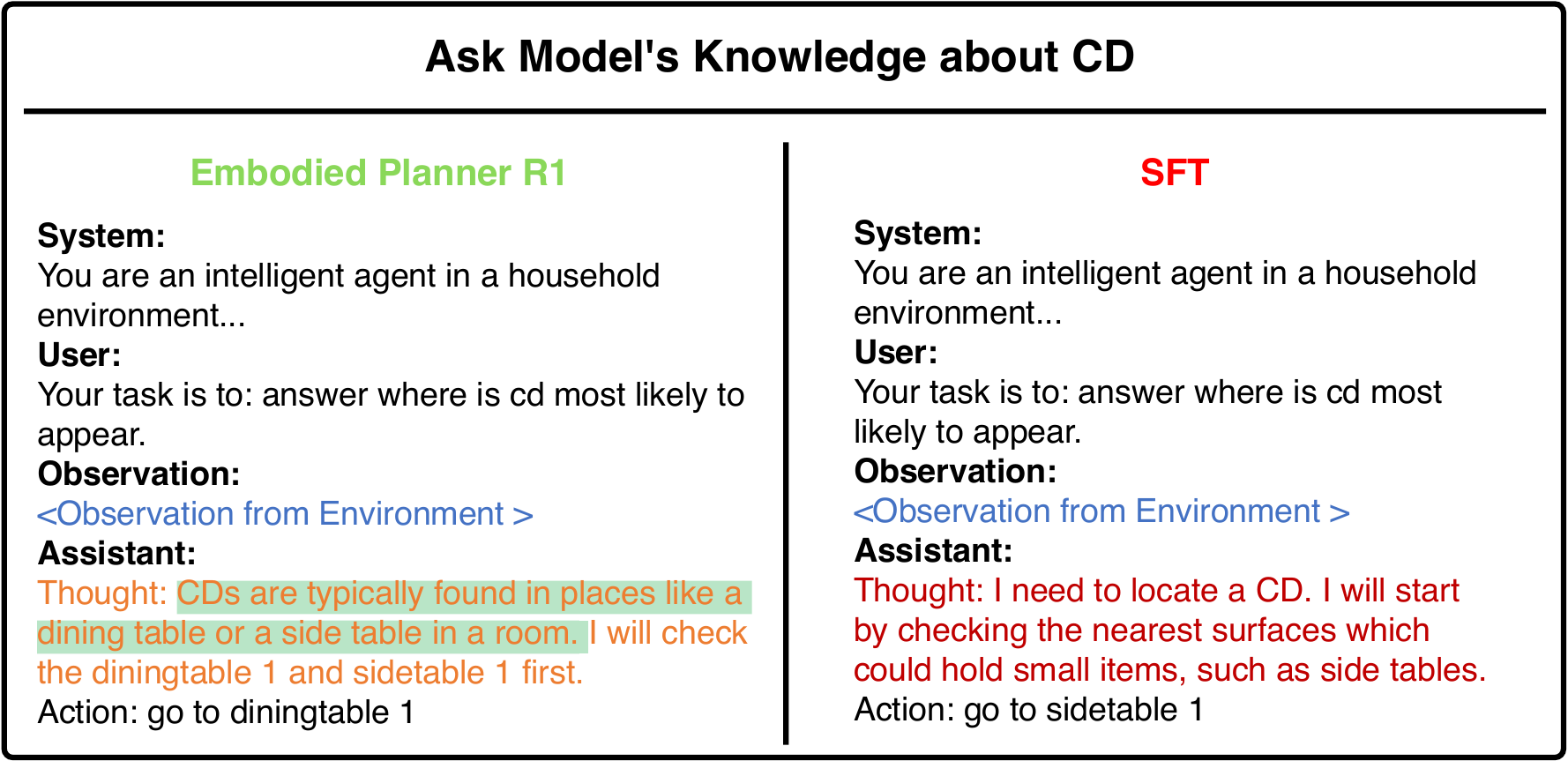}
    }
    \caption{Comparison between \framework\ and SFT when asked about the location of CD, showing SFT suffered from prior knowledge bias while \framework\ still succeeded.}
    \label{fig:case-study-ask}
\end{figure}

Beyond the model's successful generalization to both seen and unseen examples, which already indicates its understanding of environmental knowledge, we conducted a case study in ALFWorld to further compare the differences in environmental knowledge mastery between \framework\ and the SFT model.

To demonstrate that the model has acquired environmental knowledge through exploration, we identified an object, CD, that appears in numerous tasks positioned on dining tables, which contradicts common expectations where CDs would more likely be stored in drawers or on shelves.

In Figure~\ref{fig:case-study-ori}, we first maintained the input format used during training and found that although both SFT and \framework\ models made correct actions, the SFT model's thoughts did not strongly correlate with its actions. In contrast, \framework\ demonstrated awareness of CD locations in its thought process.

To further verify that \framework\ has mastered knowledge related to the environment, we modified the input format as shown in the Figure~\ref{fig:case-study-ask}, changing the task to ask the model where CDs are most likely to appear. The SFT model failed to complete this type of task, with its thoughts still reflecting the influence of prior knowledge without recognizing that CD locations in ALFWorld differ from common expectations. \framework, however, still made the correct analysis in its thoughts, proving that the model has successfully captured this implicit spatial relationship during its exploration phase.

\section{Limitations}\label{sec:limitations}

Despite our efforts, our work still has the following limitations: 

\begin{itemize}
    \item Our experiments were conducted in pure text environments, but our approach is generalizable to multimodal settings. In the future, we plan to expand our methods and extend them to the vision domain.
    \item The computational resources required for our approach may limit its accessibility for researchers with limited computing capacity. Addressing these efficiency challenges represents an important direction for future work.
\end{itemize}

\end{document}